\documentclass[runningheads]{llncs}
\usepackage{preamble}

\begin{document}
\pagestyle{headings}
\mainmatter
\def\ECCVSubNumber{1677}  

\title{Relative Pose from Deep Learned Depth and a Single Affine Correspondence} 

\author{Ivan Eichhardt\inst{1,3} \and Daniel Barath\inst{1,2}}

\authorrunning{I. Eichhardt \and D. Bar\'ath}

\institute{Machine Perception Research Laboratory, SZTAKI, Budapest, Hungary \and VRG, Faculty of Electrical Engineering, Czech Technical University in Prague \and Faculty of Informatics, University of Debrecen, Debrecen, Hungary
\email{\{eichhardt.ivan,barath.daniel\}@sztaki.mta.hu}}

\maketitle

\begin{abstract}
We propose a new approach for combining deep-learned non-metric monocular depth with affine correspondences (ACs) to estimate the relative pose of two calibrated cameras from a single correspondence. 
Considering the depth information and affine features, two new constraints on the camera pose are derived.
The proposed solver is usable within 1-point RANSAC approaches.
Thus, the processing time of the robust estimation is linear in the number of correspondences and, therefore, orders of magnitude faster than by using traditional approaches.
The proposed 1AC+D\footnote{Source code:   \faGithub\,\url{https://github.com/eivan/one-ac-pose}} solver is tested both on synthetic data and on \num{110395} publicly available real image pairs where we used an off-the-shelf monocular depth network to provide up-to-scale depth per pixel. 
The proposed 1AC+D leads to similar accuracy as traditional approaches while being significantly faster.
When solving large-scale problems, {\eg} pose-graph initialization for Structure-from-Motion (SfM) pipelines, the overhead of obtaining ACs and monocular depth is negligible compared to the speed-up gained in the pairwise geometric verification, {\ie}, relative pose estimation.
This is demonstrated on scenes from the 1DSfM dataset using a state-of-the-art global SfM algorithm.
\end{abstract}

\section{Introduction}

\begin{figure}[t]
    \centering
\begin{namedsubfigblock}{0.9\textwidth}{Vienna Cathedral}
\includegraphics[height=0.1\textheight]{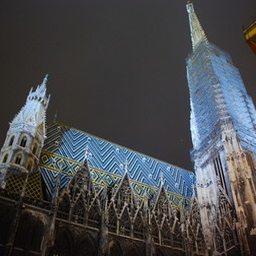}%
\includegraphics[height=0.1\textheight]{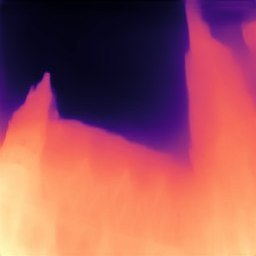}%
\includegraphics[height=0.1\textheight]{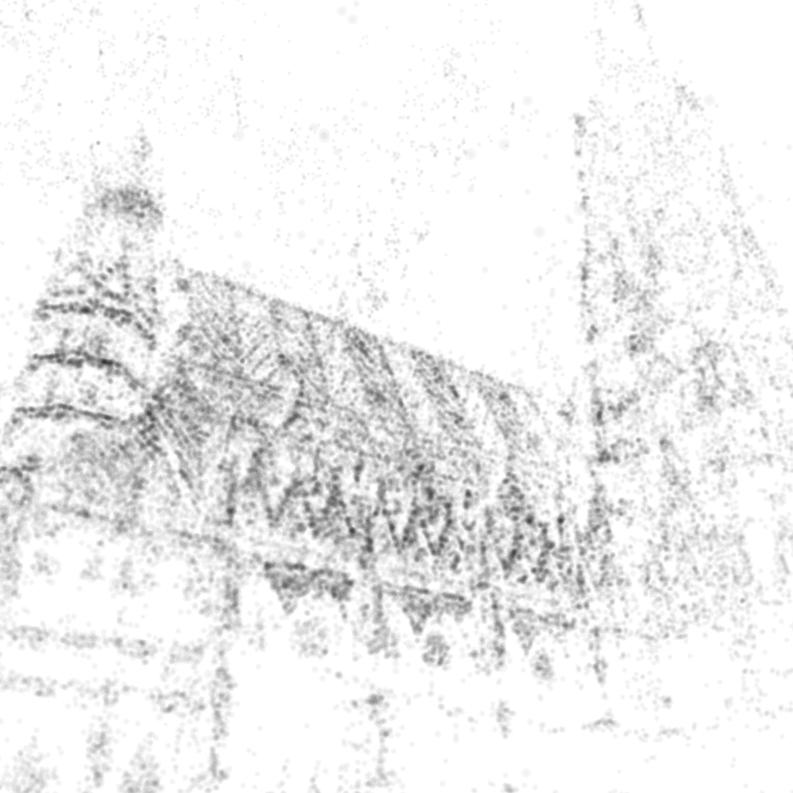}
\end{namedsubfigblock}
\begin{namedsubfigblock}{0.9\textwidth}{Madrid Metropolis}
\includegraphics[height=0.1\textheight]{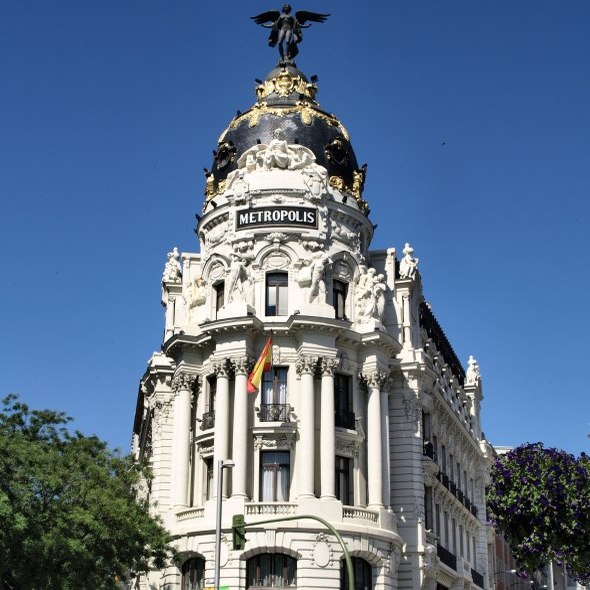}%
\includegraphics[height=0.1\textheight]{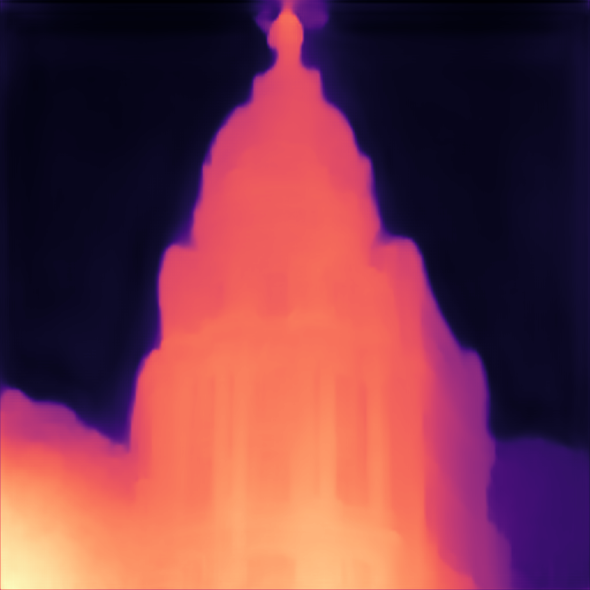}%
\includegraphics[height=0.1\textheight]{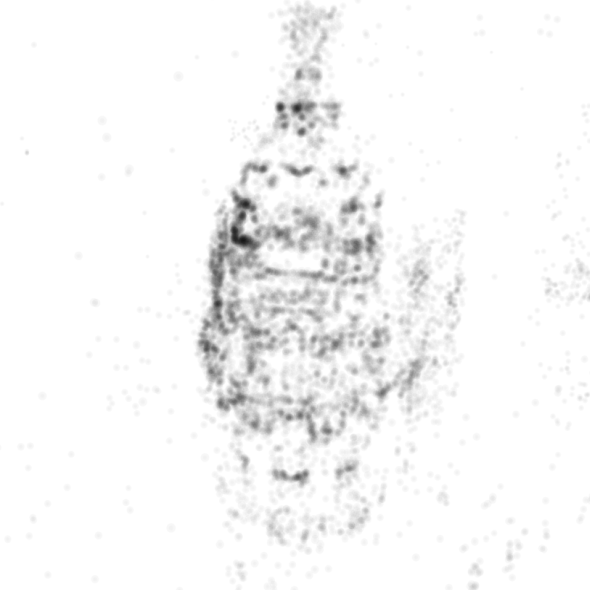}%
\end{namedsubfigblock}
    \caption{Scenes from the 1DSfM dataset~\cite{wilson_eccv2014_1dsfm}. From left to right: color image from the dataset, predicted monocular depth~\cite{Li_2018_CVPR}, and 3D reconstruction using global SfM~\cite{theia-manual}. Pose graphs were initialized by the proposed solver, 1AC+D, as written in Section~\ref{sec:experiments}.}
    \label{fig:art_color_and_depth}
\end{figure}

This paper investigates the challenges and viability of combining deep-learned monocular depth with affine correspondences (ACs) to create minimal pose solvers. 
Estimating pose is a fundamental problem in computer vision~\cite{nister2004efficient,stewenius2005minimal,stewenius2006recent,batra2007alternative,fraundorfer2010minimal,hesch2011direct,hartley2012efficient,ventura2014minimal,saurer2015minimal,nakano2016versatile,Albl_2017_CVPR,kukelova2011polynomial,Larsson_2018_CVPR,Larsson_2019_ICCV,li20134,li2006five,scaramuzza20111}, enabling reconstruction algorithms such as simultaneous localization and mapping~\cite{mur2015orb,mur2017orb} (SLAM) as well as structure-from-motion~\cite{wu2013towards,snavely2006photo,sweeney2015optimizing,schonberger2016structure} (SfM). 
Also, there are several other applications, {\eg}, in 6D object pose estimation~\cite{hodan2018bop,li2019cdpn} or in robust model fitting~\cite{fischler1981random,torr2000mlesac,chum2005matching,raguram2013usac,barath2018graph,barath2019magsac}, where the accuracy and runtime highly depends on the minimal solver applied. 
The proposed approach uses monocular depth predictions together with ACs to estimate the relative camera pose from a single correspondence.

Exploiting ACs for geometric model estimation is a nowadays popular approach with a number of solvers proposed in the recent years.
For instance, for estimating the epipolar geometry of two views, 
Bentolila and Francos~\cite{Bentolila2014} proposed a method using three correspondences interpreting the problem via conic constraints. 
Perdoch {\etal}~\cite{PerdochMC06} proposed two techniques for approximating the pose using two and three correspondences by converting each affine feature to three points and applying the standard point-based estimation techniques. 
Raposo {\etal}~\cite{Raposo2016} proposed a solution for essential matrix estimation from two correspondences. Bar\'ath {\etal}~\cite{barath2017minimal,barath2018efficient} showed that the relationship of ACs and epipolar geometry is linear and geometrically interpretable. 
Eichhardt {\etal}~\cite{eichhardt2018affine} proposed a method that uses two ACs for relative pose estimation based on general central-projective views.
Recently, Hajder {\etal}~\cite{hajder2020planar} and Guan {\etal}~\cite{guan2020minimal} proposed minimal solutions for relative
pose from a single AC when the camera is mounted on a moving vehicle.
Homographies can also be estimated from two ACs as it was first shown by Köser~\cite{koser2009geometric}.
In the case of known epipolar geometry, a single correspondence~\cite{barath2017theory} is enough to recover the parameters of the homography, or the underlying surface normal~\cite{barathEichhardtHajderTIP2019,koser2009geometric}. 
Pritts {\etal}~\cite{pritts2019minimal} used affine features for the simultaneous estimation of affine image rectification and lens distortion.

In this paper, we propose to use affine correspondences together with deep-learned priors derived directly into minimal pose solvers -- exploring new ways of utilizing these priors for pose estimation. 
We use an off-the-shelf deep monocular depth estimator~\cite{Li_2018_CVPR} to provide a `relative' (non-metric) depth for each pixel (examples are in Fig.~\ref{fig:art_color_and_depth}) and use the depth together with affine correspondences for estimating the relative pose from a single correspondence. 
Aside from the fact that predicted depth values are far from being perfect, they provide a sufficiently strong prior about the underlying scene geometry.
This helps in reducing the degrees-of-freedom of the relative pose estimation problem.
Consequently, fewer correspondences are needed to determine the pose which is highly favorable in robust estimation, {\ie}, robustly determining the camera motion when the data is contaminated by outliers and noise. 

We propose a new minimal solver which estimates the general relative camera pose, {\ie}, 3D rotation and translation, and the scale of the depth map from a single AC. 
We show that it is possible to use the predicted imperfect depth signal to robustly estimate the pose parameters.
The depth and the AC together constrain the camera geometry so that the relative motion and scale can be determined as the closed-form solution of the implied least-squares optimization problem.
The proposed new constraints are derived from general central projection and, therefore, they are valid for arbitrary pairs of central-projective views.
It is shown that the proposed solver has significantly lower computational cost, compared to state-of-the-art pose solvers.
The imperfections of the depth signal and the AC are alleviated through using the solver with a modern robust estimator~\cite{barath2018graph}, providing state-of-the-art accuracy at exceptionally low run-time.

The reduced number of data points required for the model estimation leads to linear time complexity when combined with robust estimators, {\eg} RANSAC~\cite{fischler1981random}. 
The resulting 1-point RANSAC has to check only $n$ model hypotheses (where $n$ is the point number) instead of, {\eg}, ${n}\choose{5}$ which the five-point solver implies.
This improvement in efficiency has a significant positive impact on solving large-scale problems, {\eg}, on view-graph construction ({\ie}, a number of 2-view matching and geometric verification)~\cite{snavely2006photo,snavely2008modeling,sweeney2015optimizing} which operates over thousands of image pairs.
We provide an evaluation of the proposed solver both on synthetic and real-world data.
It is demonstrated, on \num{110395} image pairs from the 1DSFM dataset~\cite{wilson_eccv2014_1dsfm}, that the proposed methods are similarly robust to image noise while being up to $2$ orders of magnitude faster, when applied within Graph-Cut RANSAC~\cite{barath2018graph}, than the traditional five-point algorithm.
Also, it is demonstrated that when using the resulting pose-graph in the global SfM pipeline~\cite{chatterjee2013efficient,wilson_eccv2014_1dsfm} as implemented in the Theia library \cite{theia-manual}, the accuracy of the obtained reconstruction is similar to when the five-point algorithm is used.

\section{Constraints from Relative Depths and Affine Frames}

Let us denote a local affine frame (LAF) as $\zaro{\ptx, \matM}$, where $\ptx\in\RR^2$ is an image point and $\matM\in\RR^{2\times2}$ is a linear transformation describing the local coordinate system of the associated image region.
The following expression maps points $\pty$ from a local coordinate system onto the image plane, in the vicinity of $\ptx$.
\begin{equation}
    \ptx\zaro{\pty} \approx \ptx + \matM \pty.
\end{equation}
LAFs typically are acquired from images by affine-covariant feature extractors~\cite{mikolajczyk2004comparison}.
In practice, they can easily be obtained by, {\eg}, the VLFeat library~\cite{vedaldi08vlfeat}.

Two LAFs $\zaro{\ptx_1, \matM_1}$ and $\zaro{\ptx_2, \matM_2}$ form an affine correspondence, namely, $\ptx_1$ and $\ptx_2$ are corresponding points and $\matM_2 \matM_1^{-1}$ is the linear transformation mapping points between the infinitesimal vicinities of $\ptx_1$ and $\ptx_2$~\cite{Eichhardt_Barath_2019_BMVC}.

\subsection{Local Affine Frames and Depth through Central Projection}

Let $q : \RR^2 \rightarrow \RR^3$ be a function that maps image coordinates to bearing vectors.
$\matR\in \text{SO}(3)$ and $\vec{t}\in\RR^3$ are the relative rotation and translation of the camera coordinate system, such that $\ptx\in\RR^2$ is the projection of $\ptX\in\RR^3$, as follows:
\begin{equation}
    \exists! \lambda : \ptX = \lambda \matR \funq{\ptx} + \vec{t}.
    \label{eq:projection}
\end{equation}
That is, a unique \emph{depth} $\lambda$ corresponds to $\ptX$, along the bearing vector $\funq{\ptx}$.

When image coordinates $\ptx$ are locally perturbed, the corresponding $\ptX$ and $\lambda$ are also expected to change as per Eq.~\eqref{eq:projection}. 
The first order approximation of the projection expresses the connection between local changes of $\ptx$ and $\ptX$, {\ie}, it is an approximate linear description how local perturbations would relate the two.
Differentiating Eq.~\eqref{eq:projection} along image coordinates $k=u,v$ results in an expression for the first order approximation of projection, as follows:
\begin{equation}
    \exists! \lambda, \partial_k\lambda : \partial_k\ptX =  \matR\zaro{\partial_k\lambda\funq{\ptx} + \lambda\partial_k\funq{\ptx} \partial_k\ptx}.
    \label{eq:projection-partial}
\end{equation}
Observe that $\vec{t}$ is eliminated from the expression above. Note that, considering the $k$-th image coordinate, $\partial_k\ptX$ is the the Jacobian of the 3D point $\ptX$, and $\partial_k\lambda$ is the Jacobian of the depth $\lambda$.
Using the differential operator $\nabla = \begin{bmatrix}\partial_u,\partial_v\end{bmatrix}$ it is more convenient to use a single compact expression, that includes the local affine frame $\matM$, as $\nabla\ptx=\begin{bmatrix}\partial_u\ptx,\partial_v\ptx\end{bmatrix}=\matM$.
\begin{equation}
    \exists! \lambda, \partial_u\lambda, \partial_v\lambda : \nabla\ptX = \matR\zaro{\funq{\ptx}\nabla\lambda + \lambda\nabla\funq{\ptx} \matM}.
    \label{eq:projection-nabla}
\end{equation}

All in all, Eqs.~\eqref{eq:projection} and \eqref{eq:projection-nabla} together describe all constraints on the 3D point $\ptX$ and on its vicinity $\nabla\ptX$, imposed by a LAF $\zaro{\ptx, \matM}$.

\subsection{Affine Correspondence and Depth Constraining Camera Pose}

Assume two views observing $\ptX$ under corresponding LAFs $\zaro{\ptx_1, \matM_1}$ and $\zaro{\ptx_2, \matM_2}$.
Without the loss of generality, we define the coordinate system of the first view as the identity -- thus putting it in the origin -- while that of the second one is described by the relative rotation and translation, $\matR$ and $\vec{t}$, respectively.
In this two-view system, Eqs.~\eqref{eq:projection} and \eqref{eq:projection-nabla} lead to two new constraints on the camera pose, depth and local affine frames, as follows: 
\begin{equation}
    \begin{aligned}
        \underbrace{\lambda_1 \funqy{1}{\ptx_1}}_\vec{a} &= \matR\underbrace{\lambda_2 \funqy{2}{\ptx_2}}_\vec{b} + \vec{t},\\
        \underbrace{\funqy{1}{\ptx_1}\nabla\lambda_1 + \lambda_1\nabla\funqy{1}{\ptx_1} \matM_1}_\vec{A} &= \matR\underbrace{\zaro{\funqy{2}{\ptx_2}\nabla\lambda_2 + \lambda_2\nabla\funqy{2}{\ptx_2} \matM_2}}_\vec{B},
    \end{aligned}
    \label{eq:constraints-nabla}
\end{equation}
where the projection functions $\funqy{1}{\ptx_1}$ and $\funqy{2}{\ptx_2}$ assign bearing vectors to image points $\ptx_1$ and $\ptx_2$ of the first and second view, respectively.
The first equation comes from Eq.~\eqref{eq:projection} through the point correspondence, and the second one from Eq.~\eqref{eq:projection-nabla} through the AC, by equating $\ptX$ and $\nabla\ptX$, respectively.
Note that depths $\lambda_1$, $\lambda_2$ and their Jacobians $\nabla\lambda_1$, $\nabla\lambda_2$ are intrinsic to each view.
Also observe that 3D point $\ptX$ and its Jacobian $\nabla\ptX$ are now eliminated from these constraints.

In the rest of the paper, we are resorting to the more compact notations using $\vec{a}, \vec{b}, \vec{A}$ and $\vec{B}$, as highlighted above, {\ie}, the two lines of Eq.~\eqref{eq:constraints-nabla} take the simplified forms of $\vec{a} = \matR \vec{b} + \vec{t}$ and $\vec{A} = \matR \vec{B}$, respectively.

\subsubsection{Relative depth.}
In this paper, as monocular views are used to provide relative depth predictions, $\lambda_1$ and $\lambda_2$ are only known up to a common scale $\Lambda$, so that Eq.~\eqref{eq:constraints-nabla}, with the simplified notation, is modified as follows:
\begin{equation}
    \begin{aligned}
        \vec{a} &= \Lambda \matR \vec{b} + \vec{t},\quad
        \vec{A} &= \Lambda \matR \vec{B}.
    \end{aligned}
    \label{eq:constraints-nabla-relativedepth}
\end{equation}
These constraints describe the relationship of relative camera pose and ACs in the case of known relative depth.

\section{Relative Pose and Scale from a Single Correspondence}

Given a single affine correspondence, the optimal estimate for the relative pose and scale is given as the solution of the following optimization problem.
\begin{equation}
    \min_{\matR, \vec{t}, \Lambda} \underbrace{\frac{1}{2}\zaroVert{\vec{a} - \zaro{\Lambda \matR \vec{b} + \vec{t}}}_2^2 + \frac{1}{2}\zaroVert{\vec{A} - \Lambda \matR \vec{B}}_{\text{F}}^2}_{f\zaro{\Lambda, \matR, \vec{t}}}
    \label{eq:opt-problem}
\end{equation}
To solve the problem, the first order optimality conditions have to be investigated.
At optimality, differentiating $f\zaro{\Lambda, \matR, \vec{t}}$ by $\vec{t}$ gives:
\begin{equation}
    \nabla_\vec{t} f\zaro{\Lambda, \matR, \vec{t}} = \vec{a} - \Lambda \matR \vec{b} - \vec{t} \stackrel{!}{=}\vec{0},
\end{equation}
that is, $\vec{t}$ can only be determined once the rest of the unknowns, $\matR$ and $\Lambda$, are determined.
This also means that above optimization problem can be set free of the translation, by performing the following substitution  $\vec{t} \leftarrow \Lambda\matR\vec{b} - \vec{a}$.

The optimization problem, where the translation is replaced as previously described, only contains the rotation and scale as unknowns, as follows:
\begin{equation}
    \min_{\matR, \Lambda} \underbrace{\frac{1}{2}\zaroVert{\vec{A} - \Lambda \matR \vec{B}}_{\text{F}}^2}_{g\zaro{\matR, \Lambda}}.
    \label{eq:opt-problem-subst-for-t}
\end{equation}
Below, different approaches for determining the unknown scale and rotation are described, solving Eq.~\eqref{eq:opt-problem-subst-for-t} exactly or, in some manner, approximately.

\subsubsection{1AC+D~(Umeyama): An SVD solution.}
In the least-squares sense, the optimal rotation and scale satisfying Eq.~\eqref{eq:opt-problem-subst-for-t} can be acquired by the singular value decomposition of the covariance matrix of $\vec{A}$ and $\vec{B}$, as follows:
\begin{equation}
    \vec{U}\vec{S}\vec{V}^\transpose = \vec{A}\vec{B}^\transpose.
\end{equation}
Using these matrices, the optimal rotation and scale are expressed as
\begin{equation}
    \begin{aligned}
        \matR &= \vec{U} \vec{D} \vec{V}^\transpose,\quad
        \Lambda &= \frac{\tr\zaro{\vec{S}\vec{D}}}{\tr\zaro{\vec{B}^\transpose\vec{B}}},
    \end{aligned}    
\end{equation}
where $\vec{D}$ is a diagonal matrix with its diagonal being $\begin{bmatrix}1, 1, \det(\vec{U}\vec{V}^\transpose)\end{bmatrix}$, to constrain $\det{\matR}=1$.
Given $\matR$ and $\Lambda$, the translation is expressed as $\vec{t} = \vec{a} - \Lambda \matR \vec{b}$.

This approach was greatly motivated by Umeyama's method~\cite{umeyama1991least}, where a similar principle is used for solving point cloud alignment.

\subsubsection{1AC+D~(Proposed): Approximate relative orientation solution.}
Matrices $\vec{A}$ and $\vec{B}$, together with their 1D left-nullspaces, define two non-orthogonal coordinate systems.
By orthogonalizing the respective basis vectors, one can construct matrices $\matR_\vec{A}$ and $\matR_\vec{B}$, corresponding to $\vec{A}$ and $\vec{B}$.
The relative rotation between these two coordinate systems can be written as follows:
\begin{equation}
    \matR = \matR_\vec{A}^\transpose \matR_\vec{B}^{\,}.
    \label{eq:1acprop_R}
\end{equation}
$\matR_\vec{A}$ and $\matR_\vec{B}$ can be determined, {\eg}, using Gram-Schmidt orthonormalization~\cite{solomon2015numerical}.
For our special case, a faster approach is shown in Alg.~\ref{alg:1ac_algorithm}.

Using this approach, the solution is biased towards the first columns of $\vec{A}$ and $\vec{B}$, providing perfect alignment of those two axes.
In our experiments, this was rather a favorable property than an issue. 
The first axis of a LAF represents the orientation of the feature while the second one specifies the affine shape, as long as the magnitude of $\nabla\lambda_i$ is negligible compared to $\lambda_i$, which is usually the case.
The orientation of a LAF is usually more reliable than its shape.

As $\matR$ is now known and $\vec{t}$ has been ruled out, $\Lambda$ is to be determined.
Differentiating $g\zaro{\matR, \Lambda}$ by $\Lambda$ gives:
\begin{equation}
    \nabla_\Lambda g\zaro{\matR, \Lambda} = \tr\zaro{\vec{A}^\transpose \matR \vec{B}} + \Lambda \tr\zaro{\vec{B}^\transpose\vec{B}},
\end{equation}
that is, once $\matR$ is known, $\Lambda$ can be determined as follows:
\begin{equation}
    \Lambda = -\frac{\tr\zaro{\vec{A}^\transpose \matR \vec{B}}}{\tr\zaro{\vec{B}^\transpose\vec{B}}}.
    \label{eq:1acprop_lambda}
\end{equation}
Finally, the translation parameters are expressed as $\vec{t} = \vec{a} - \Lambda \matR \vec{b}$.

The complete algorithm is shown in Alg.~\ref{alg:1ac_algorithm}.
Note that, although, we have tested various approaches to computing the relative rotation between the frames $\vec{A}$ and $\vec{B}$, such as the above described SVD approach or the Gram-Schmidt process~\cite{solomon2015numerical}, the one introduced in Alg.~\ref{alg:1ac_algorithm} proved to be the fastest with no noticeable deterioration in accuracy.

\begin{table}[t]
\center
\caption{The theoretical computational complexity of the solvers used in RANSAC. The reported properties are: the number of operations of each solver (steps; $1$st row); the computational complexity of one estimation (1 iter; $2$nd); the number of correspondences required for the estimation ($m$; $3$rd); possible outlier ratios ($1 - \mu$; $4$th); the number of iterations needed for RANSAC with the required confidence set to $\num{0.99}$ (\# iters; $5$th); and computational complexity of the full procedure (\# comps; $6$th).
The major operations are: singular value decomposition (SVD), eigenvalue decomposition (EIG), LU factorization (LU), and QR decomposition (QR).}
\label{tab:computational_complexity}
\setlength\aboverulesep{0pt} 
\setlength\belowrulesep{0pt}
\resizebox{0.99\textwidth}{!}{%
\begin{tabular}{r|ccc|ccc||ccc|ccc}
\toprule
\rowcolor{black!10} 
 &
  \multicolumn{3}{c|}{\cellcolor{black!10}point-based} &
  \multicolumn{3}{c||}{\cellcolor{black!10}AC-based} &
  \multicolumn{6}{c}{\cellcolor{black!10}AC+depth} \\ \midrule
\rowcolor{black!10} 
 &
  \multicolumn{3}{c|}{\cellcolor{black!10}5PT~\cite{stewenius2006recent}} &
  \multicolumn{3}{c||}{\cellcolor{black!10}2AC~\cite{barath2018efficient}} &
  \multicolumn{3}{c|}{\cellcolor{black!10}1AC+D~(Umeyama)} &
  \multicolumn{3}{c}{\cellcolor{black!10}1AC+D~(Proposed)} \\ \midrule
\setlength\aboverulesep{3pt} 
\setlength\belowrulesep{3pt}
steps &
  \multicolumn{3}{c|}{\begin{tabular}[c]{@{}c@{}}5×9 SVD + 10×20 LU\\ + 10×10 EIG\end{tabular}} &
  \multicolumn{3}{c||}{\begin{tabular}[c]{@{}c@{}}6×9 SVD + 10×9 QR\end{tabular}} &
  \multicolumn{3}{c|}{\begin{tabular}[c]{@{}c@{}}3×3 SVD\\ + cov.\ + trans.\end{tabular}} &
  \multicolumn{3}{c}{\begin{tabular}[c]{@{}c@{}}$4\times\zaro{\text{cross}+\text{norm}}$\\  + trans.\ + etc.\end{tabular}} \\ \midrule
1 iter &
  \multicolumn{3}{c|}{\begin{tabular}[c]{@{}c@{}}$5^2 * 9 + 10^2 * 20 + 10^2$\\ $=2325$\end{tabular}} &
  \multicolumn{3}{c||}{\begin{tabular}[c]{@{}c@{}}$6^2 * 9 + 10^3 + 10^2$\\ $=1424$\end{tabular}} &
  \multicolumn{3}{c|}{\begin{tabular}[c]{@{}c@{}}$11 * 3^3 + 27 + 21$ \\ $=345$\end{tabular}} &
  \multicolumn{3}{c}{\begin{tabular}[c]{@{}c@{}}$72 + 21 + 40$\\ $=133$\end{tabular}} \\ \midrule
$m$ &
  \multicolumn{3}{c|}{5} &
  \multicolumn{3}{c||}{2} &
  \multicolumn{3}{c|}{1} &
  \multicolumn{3}{c}{1} \\ \midrule
\multicolumn{1}{c|}{\begin{tabular}[c]{@{}r@{}} 1-$\mu$\\ \# iters\\ \# comps\end{tabular}} &
  \multicolumn{3}{c|}{\begin{tabular}[c]{@{}ccc@{}}0.50 & 0.75 & 0.90\\ \num{145} & \num{4713} & $\sim10^6$ \\ $\sim10^5$ & $\sim10^7$ & $\sim10^9$ \end{tabular}} &
  \multicolumn{3}{c||}{\begin{tabular}[c]{@{}ccc@{}}0.50 & 0.75 & 0.90\\ \num{16} & \num{71} & \num{458} \\ $\sim10^4$ & $\sim10^5$ & $\sim10^5$  \end{tabular}} &
  \multicolumn{3}{c|}{\begin{tabular}[c]{@{}ccc@{}}0.50 & 0.75 & 0.90\\ \num{7} & \num{16} & \num{44} \\ $\sim10^3$ & $\sim10^3$ & $\sim10^4$  \end{tabular}} &
  \multicolumn{3}{c}{\begin{tabular}[c]{@{}ccc@{}}0.50 & 0.75 & 0.90\\ \num{7} & \num{16} & \num{44} \\ $\sim10^2$ & $\sim10^3$ & $\sim10^3$  \end{tabular}} \\ \bottomrule
\end{tabular}%
}%
\end{table}

\begin{algorithm}
    \caption{The 1AC+D~(Proposed) algorithm for relative pose computation.}
    \label{alg:1ac_algorithm}
    \begin{algorithmic}[1]
        \Procedure{1AC+D~(Proposed)}{$\vec{a},\vec{b},\vec{A},\vec{B}$} \Comment{Computes the relative camera pose.}
            \State $\matR_\vec{A} \gets \Call{orthonorm}{\vec{A}}$
            \State $\matR_\vec{B}^{\,} \gets \Call{orthonorm}{\vec{B}}$
            \State $\matR \gets \matR_\vec{A}^\transpose \matR_\vec{B}^{\,}$ \Comment{Applying Eq.~\eqref{eq:1acprop_R}.}
            \State $\Lambda \gets -{\tr\zaro{\vec{A}^\transpose \matR \vec{B}}} / {\tr\zaro{\vec{B}^\transpose\vec{B}}}$ \Comment{Applying Eq.~\eqref{eq:1acprop_lambda}.}
            \State $\vec{t} \gets \vec{a}-\Lambda\matR\vec{b}$
            \State \textbf{return} $\matR, \vec{t}, \Lambda$
        \EndProcedure
        \Function{orthonorm}{$\vec{Y}$} \Comment{Quick orthonormalization of $\vec{Y}$.}
            \State $\vec{r}_x \gets \mathrm{normalize}\zaro{\vec{Y}_{\zaro{:,1}} \times \vec{Y}_{\zaro{:,2}}}$ \Comment{$\vec{r}_x$ is a normal to the underlying plane.}
            \State $\vec{r}_z \gets \mathrm{normalize}\zaro{\vec{Y}_{\zaro{:,2}}}$
            \State \textbf{return} $\begin{bmatrix}\vec{r}_x, & \vec{r}_z \times \vec{r}_x, & \vec{r}_z\end{bmatrix}$ \Comment{Return a $3\times3$ rotation matrix.}
        \EndFunction
    \end{algorithmic}
\end{algorithm}

\section{Experimental results}

In this section, experiments and complexity analyses compare the performance of the two versions of the proposed 1AC+D solver, {\ie}, 1AC+D~(Umeyama) and 1AC+D~(Proposed), 2AC~\cite{barath2018efficient} and 5PT~\cite{stewenius2006recent} methods for relative pose estimation.

\subsection{Processing time}

In Table~\ref{tab:computational_complexity}, we compare the computational complexity of state-of-the-art pose solvers used in our evaluation.
The first row consists of the major steps of each solver. For instance, $5 \times 9$ SVD + $10 \times 20$ EIG means that the major steps are: the SVD decomposition of a $5 \times 9$ matrix and eigendecomposition of a $10 \times 20$ matrix. 
In the second row, the implied computational complexities are summed. 
In the third one, the number of correspondences required for the solvers are written. 
The fourth row lists example outlier ratios. 
In the fifth one, the theoretical numbers of iterations of RANSAC~\cite{hartley2003multiple} are written with confidence set to $\num{0.99}$.
The last row reports the total complexity, {\ie}, the complexity of one iteration multiplied by the number of RANSAC iterations. 

The proposed methods have significantly smaller computational requirements than the five-point solver, 5PT, or the affine correspondence-based 2AC method \cite{barath2018efficient} when included in RANSAC.
Note that while the reported values for the 1AC+D methods are the actual FLOPs, for 5PT and 2AC, it is in practice about an order of magnitude higher due to the iterative manner of various linear algebra operations, {\eg}, SVD.
%
\begin{figure}[t]
    \centering
    \includegraphics[width=0.5\textwidth]{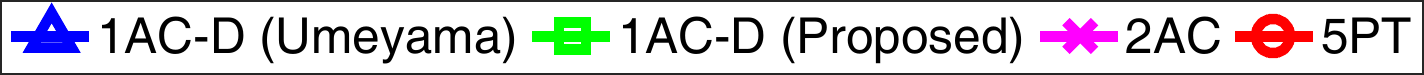}

    \begin{resizeblock}{0.40\linewidth}%
	    \includegraphics[width=1.0\textwidth,trim={0.5 0 1.0cm 0},clip]{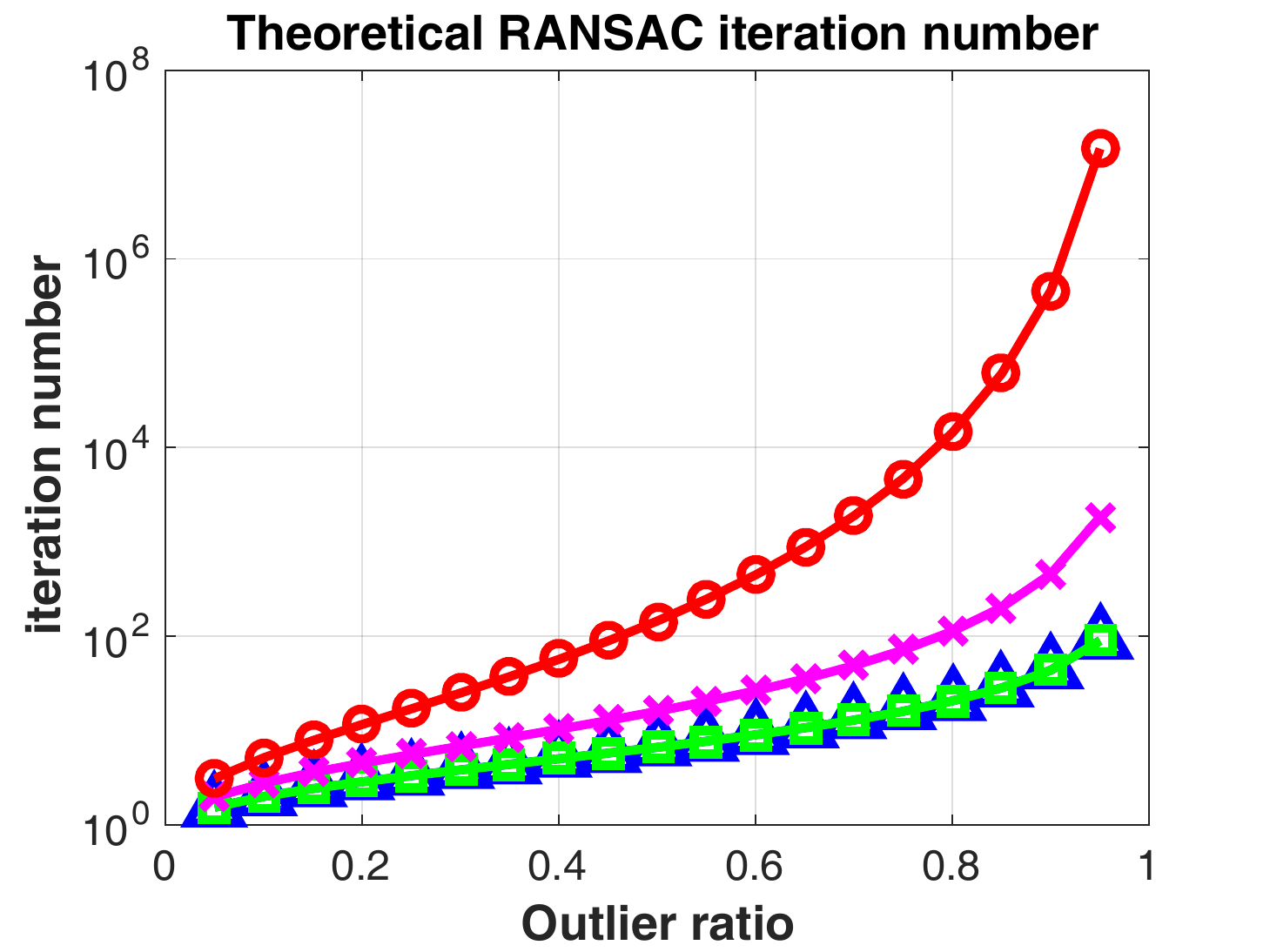}%
        \label{fig:theoretical_ransac_iterations}%
	\end{resizeblock}\quad
	\begin{resizeblock}{0.40\linewidth}%
	    \includegraphics[width=1.0\textwidth,trim={0.5 0 1.0cm 0},clip]{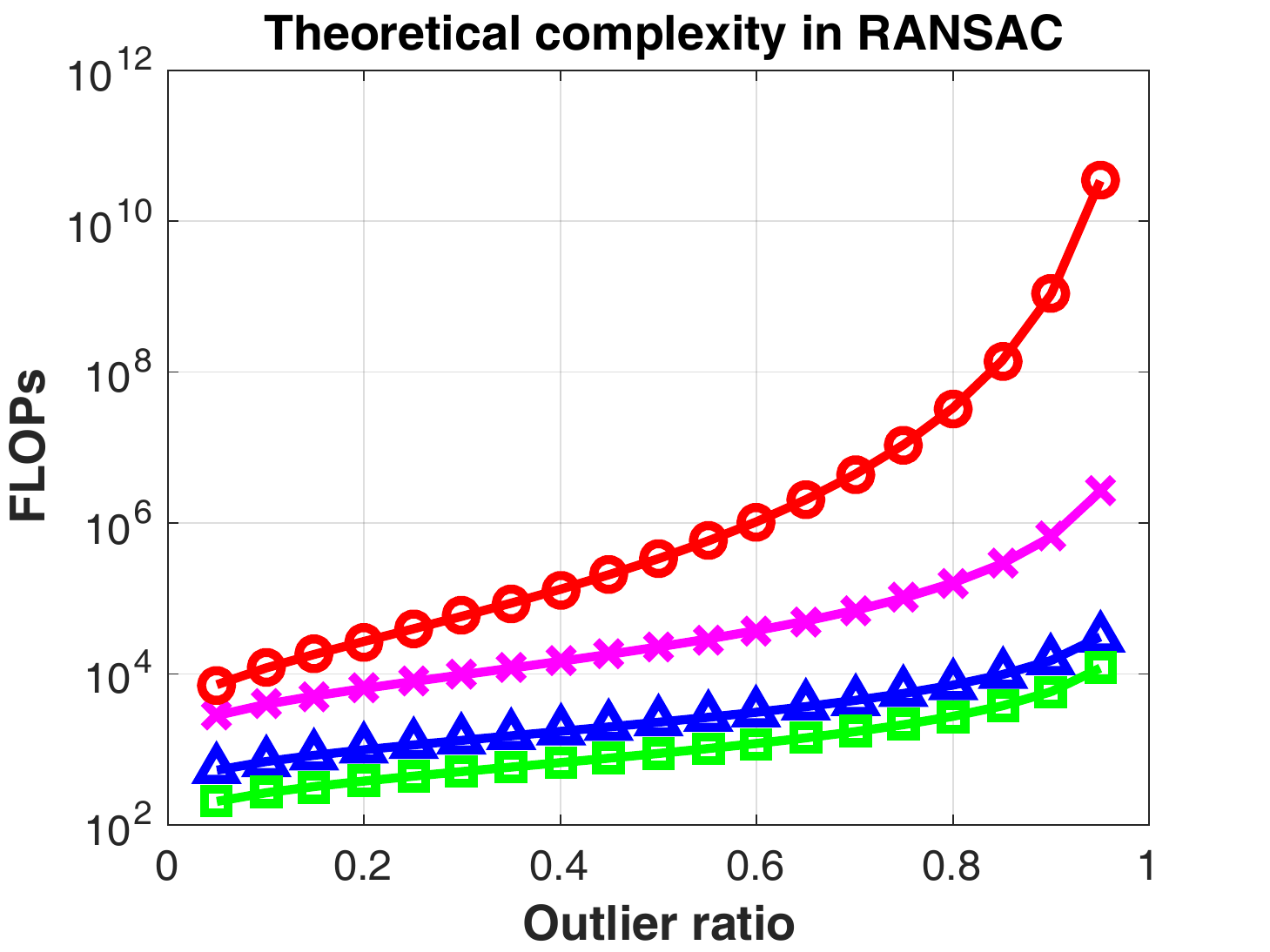}%
        \label{fig:theoretical_ransac_flops}%
	\end{resizeblock}%
    \caption{\textit{Left:} the theoretical number of RANSAC iterations -- calculated as in~\cite{hartley2003multiple}. \textit{Right:} the number of floating point operations (horizontal axes) plotted as the function of the outlier ratio (vertical), displayed on logarithmic scales.}
    \label{fig:theoretical_ransac}%
\end{figure}
%
In Fig.~\ref{fig:theoretical_ransac}, the theoretical number of RANSAC iterations -- calculated as in~\cite{hartley2003multiple} -- and the number of floating point operations are plotted as the function of the outlier ratio (horizontal axes), displayed on logarithmic scales (vertical axes). 
The proposed solvers lead to fewer iterations compared to the solvers solely based on point or affine correspondences. 

In summary of the above analysis of the computational complexity, the proposed approach speeds up the robust estimation procedure in three ways. 
First, it requires at least an order of magnitude fewer operations to complete a single iteration of RANSAC than by using the five-point algorithm.
Second, it leads to significantly fewer iterations due to the fact that 1AC+D takes only a single correspondence to propose a solution to pose estimation.
Third, 1AC+D returns a single solution from a minimal sample, in contrast to the five-point algorithm. Therefore, each RANSAC iteration requires the validation of a single model. 

\subsection{Synthetic evaluation}

The synthetic evaluation was carried out in a setup consisting of two cameras with randomized poses represented by rotation $\matR_i\in \text{SO}\zaro{3}$ and translation $\vec{t}_i\in\RR^{3}$ $(i=1,2)$.
The cameras were placed around the origin at a distance sampled from $\left[1.0, 2.0\right]$, oriented towards a random point sampled from $\left[-0.5, 0.5\right]^3$.
Both cameras had a common intrinsic camera matrix $\vec{K}$ with focal length $f = \num{600}$ and principal point $\left[300, 300\right]$, representing pinhole projection $\vec{\pi}:\RR^3\rightarrow\RR^2$.
For generating LAFs $\zaro{\ptx_i,\matM_i}$, depths $\lambda_i$ and their derivatives $\nabla\lambda_i$, randomized 3D points $\ptX\sim\mathcal{N}\zaro{\vec{0},\vec{I}_{3\times3}}$ with corresponding normals $\vec{n}\in\RR^3$, $\zaroVert{\vec{n}}_2=1$ were projected into the two image planes using Eqs.~\eqref{eq:howsyntheticdataisgenerated_affine} and~\eqref{eq:howsyntheticdataisgenerated_depth}, for $i=1,2$.
\begin{align}
&\ptx_i = \vec{\pi} \zaro{\matR_i \ptX + \vec{t}_i}, &\matM_i& = \nabla\vec{\pi}\zaro{\matR_i \ptX + \vec{t}_i} \matR_i \nabla\ptX,\label{eq:howsyntheticdataisgenerated_affine}\\
&\lambda_i = \left.\matR_i\right|_{\zaro{3,:}} \ptX + \left.\vec{t}_i\right|_{\zaro{3}}, &\nabla\lambda_i& = \left.\matR_i\right|_{\zaro{3,:}} \nabla\ptX,\label{eq:howsyntheticdataisgenerated_depth}
\end{align}
where $\left.\matR_i\right|_{\zaro{3,:}}$ is the 3rd row of $\matR_i$, $\nabla\ptX=\mathrm{nulls}\zaro{\vec{n}}$ simulates the local frame of the surface as the nullspace of the surface normal $\vec{n}$.
Note that \cite{eichhardt2018affine} proposed the approach for computing local frames $\vec{M}_i$, {\ie}, Eq.~\eqref{eq:howsyntheticdataisgenerated_affine}.
Their synthetic experiment are also similar in concept, but contrast to them, we also had to account for the depth in Eq.~\eqref{eq:howsyntheticdataisgenerated_depth}, and we used the nullspace of ${\vec{n}}$ to express the Jacobian $\nabla\ptX$.
The 2nd part of Eq.~\eqref{eq:howsyntheticdataisgenerated_depth}, the expression for the Jacobian of the depth, $\nabla\lambda_i$, was derived from the 1st part by differentiation.
In this setup $\lambda_i$ is the projective depth, a key element of perspective projection.
Note that the depth and its derivatives are different for various other camera models.
Finally, zero-mean Gaussian-noise was added to the coordinates/components of $\ptx_i$, $\matM_i$, $\lambda_i$ and $\nabla\lambda_i$, with $\sigma$, $\sigma_\matM$, $\sigma_\lambda$ and $\sigma_{\nabla\lambda}$ standard deviation (STD), respectively. 

The reported errors for rotation and translation both were measured in degrees (\SI{}{\degree}) in this evaluation. The rotation error was calculated as follows:
$\epsilon_\matR = \cos^{-1}\left(\frac{1}{2} \left(\tr{(\hat{\matR} \matR^\transpose)} - 1\right)\right)$,
where $\hat{\matR}$ is the measured and $\matR$ is the ground truth rotation matrix.
Translation error $\epsilon_\textbf{t}$ is the angular difference between the estimated and ground truth translations.
For some of the tests, we also show the avg. Sampson error~\cite{hartley2003multiple} of the implied essential matrix.

\begin{figure*}[t]
\centering
\includegraphics[width=0.5\textwidth]{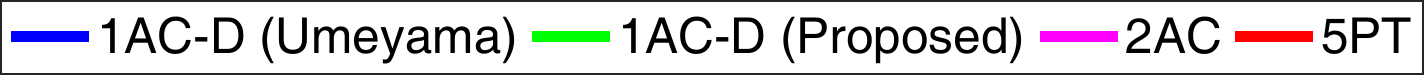}\\
	\begin{subfigblock}{0.30\linewidth}
	    \includegraphics[width=1.0\textwidth,trim={0.5 0 1.0cm 0},clip]{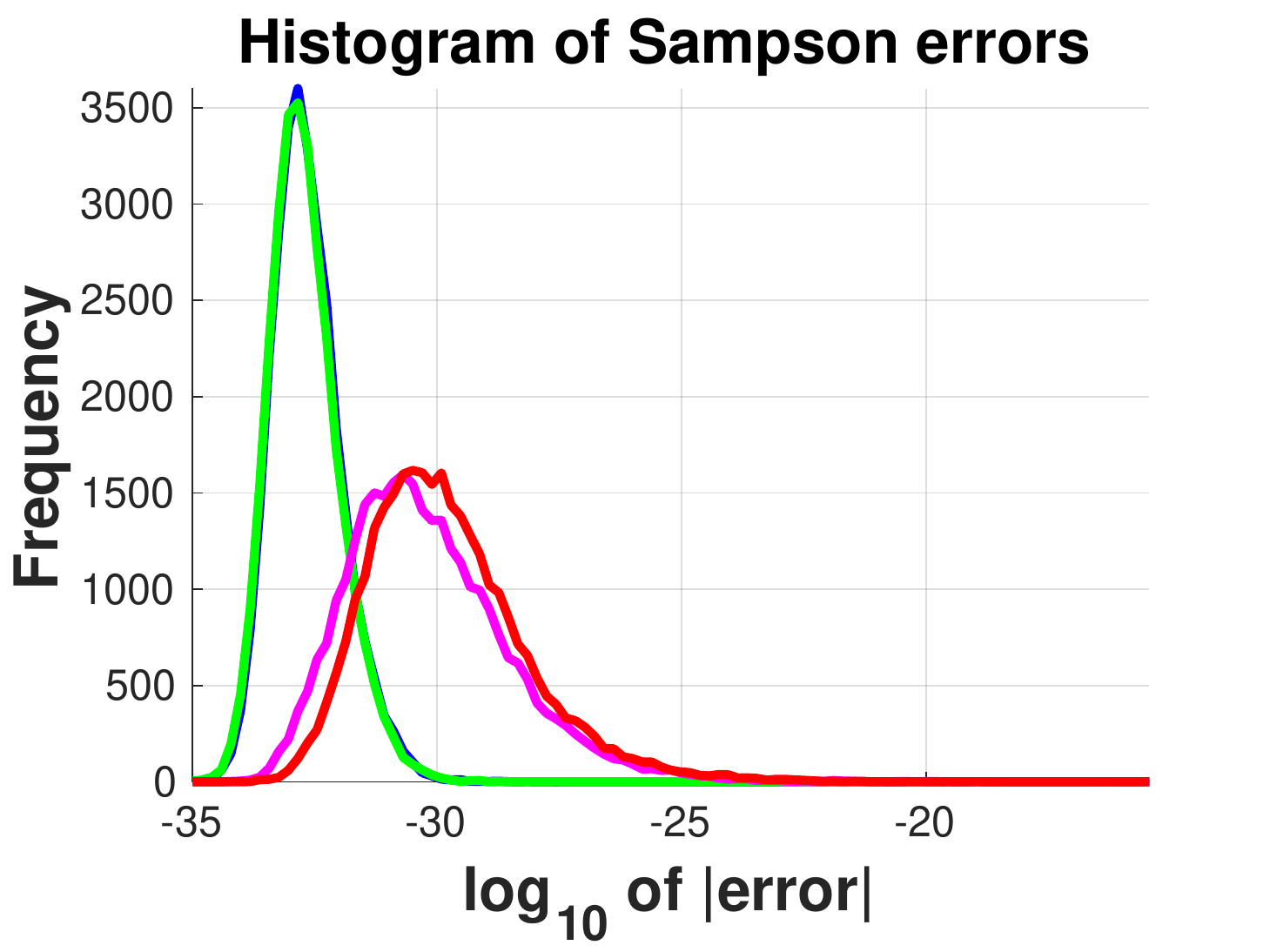}
        \label{fig:stability_a}
	\end{subfigblock}\hfill
	\begin{subfigblock}{0.30\linewidth}
	    \includegraphics[width=1.0\textwidth,trim={0.5 0 1.0cm 0},clip]{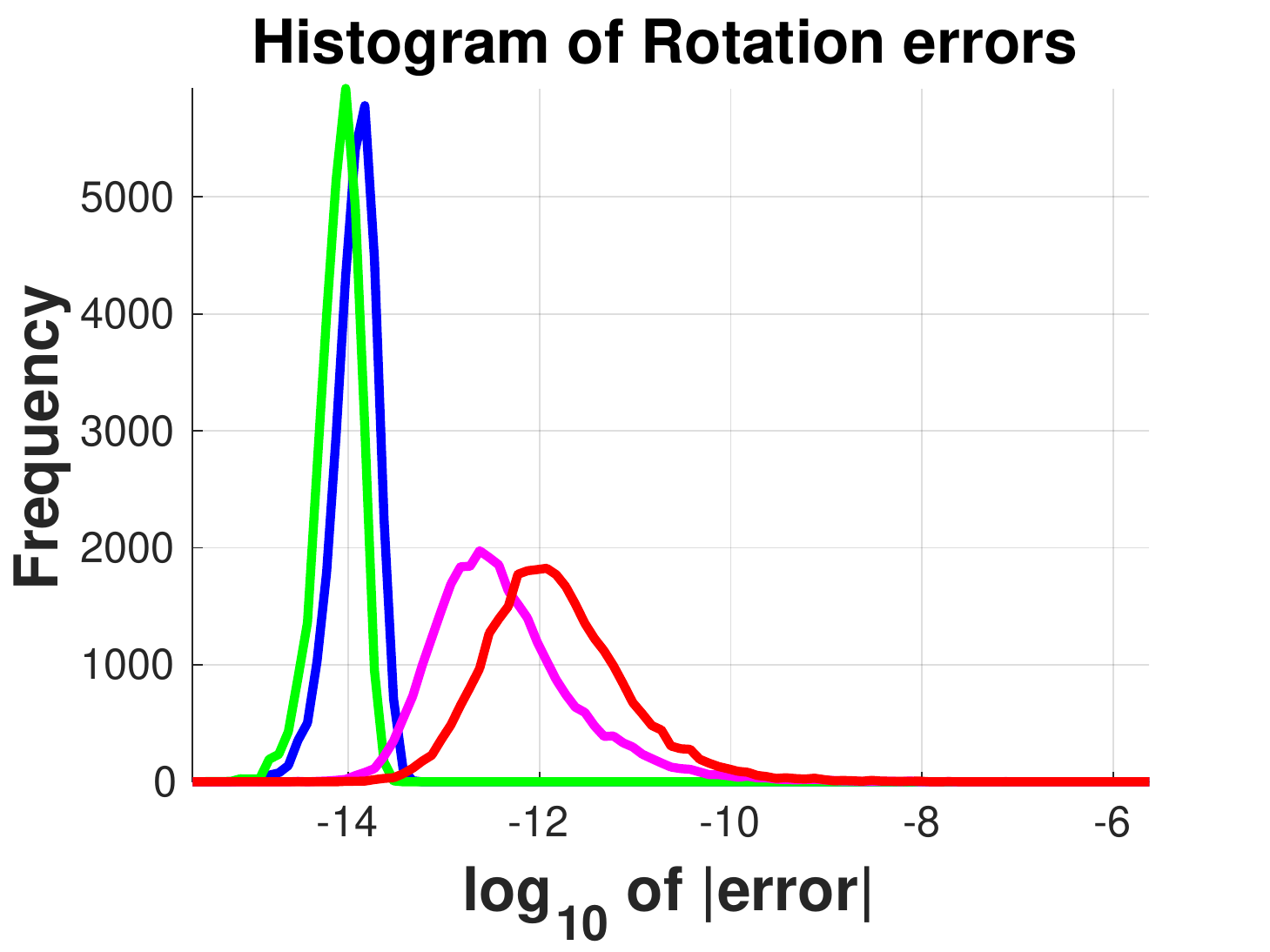}
        \label{fig:stability_b}
	\end{subfigblock}\hfill
	\begin{subfigblock}{0.30\linewidth}
	    \includegraphics[width=1.0\textwidth,trim={0.5 0 1.0cm 0},clip]{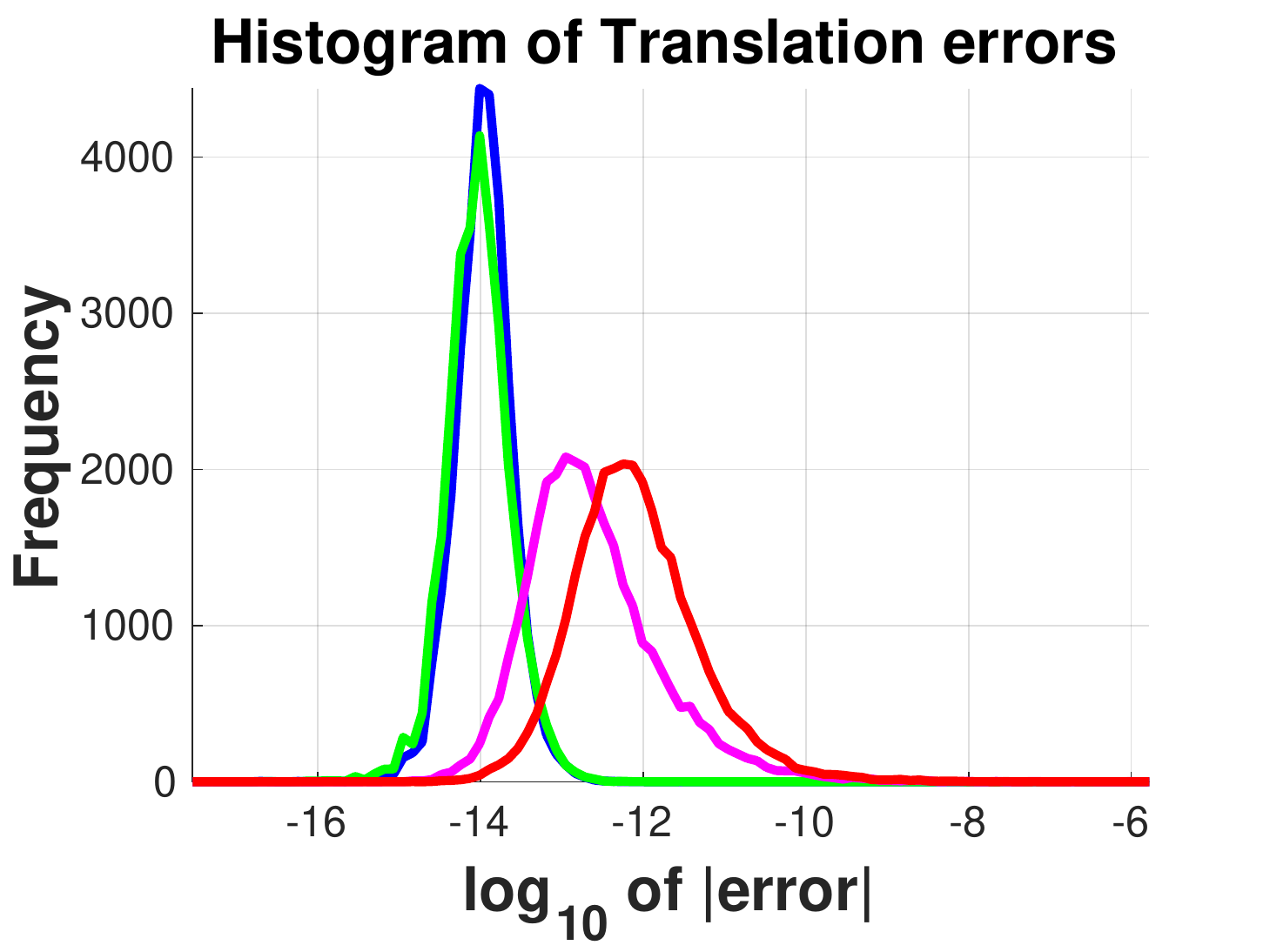}
        \label{fig:stability_c}
	\end{subfigblock}
    \caption{Stability study of four different relative pose solvers: two versions of the proposed 1AC+D method, 1AC+D~(Umeyama) and 1AC+D~(Proposed), 2AC~\cite{barath2018efficient} solver using two affine correspondences and the five point solver 5PT~\cite{stewenius2006recent}. The three plots show the distribution of various errors on logarithmic scales as histograms, namely {(a)}~Sampson error, {(b)}~rotation  and {(c)}~translation errors in degrees.}
\end{figure*}

\noindent \textbf{Solver stability.}
We randomly generated \num{30000} instances of the above described synthetic setup, except for adding any noise to the LAFs and depths, in order to evaluate the stability of the various solver in a noise-free environment.
Figs.~\ref{fig:stability_a}, \ref{fig:stability_b} and \ref{fig:stability_c} show the distribution of the Sampson, rotation and translation errors on logarithmic scales.
The figures show that the proposed method is one of the best performers given that its distribution of errors is shifted lower compared to the other methods.
However, all methods are quite stable having no peaks on the right side of the figures.

\begin{figure*}[t]
\centering
\includegraphics[width=0.5\textwidth]{assets/noisebar_markers.pdf}\\
	\begin{subfigblock}{0.30\linewidth}
	    \includegraphics[trim={0.5cm 0 1cm 0},clip]{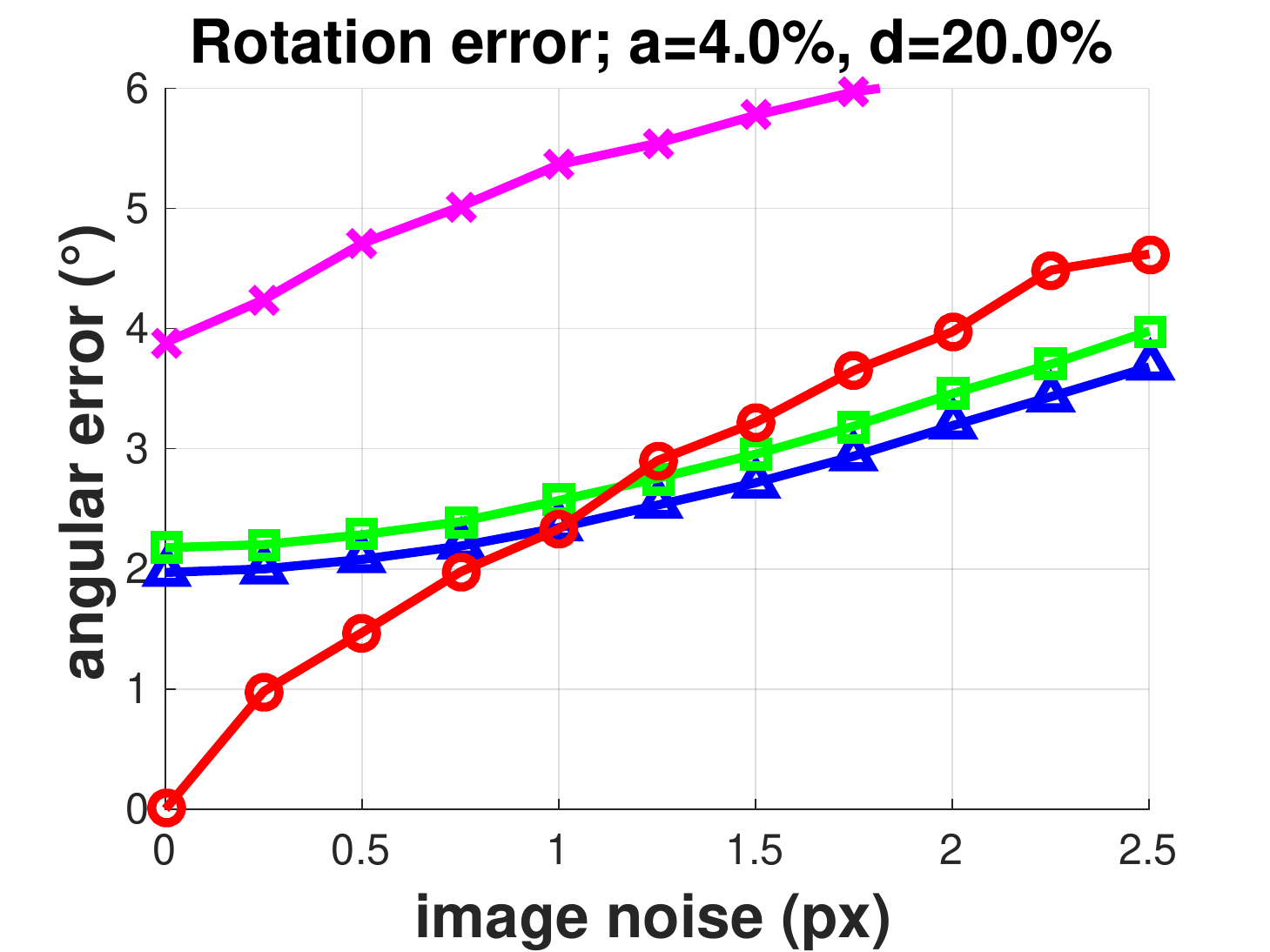}
        \label{fig:noisy_a}
	\end{subfigblock}\hfill
	\begin{subfigblock}{0.30\linewidth}
	    \includegraphics[trim={0.5cm 0 1cm 0},clip]{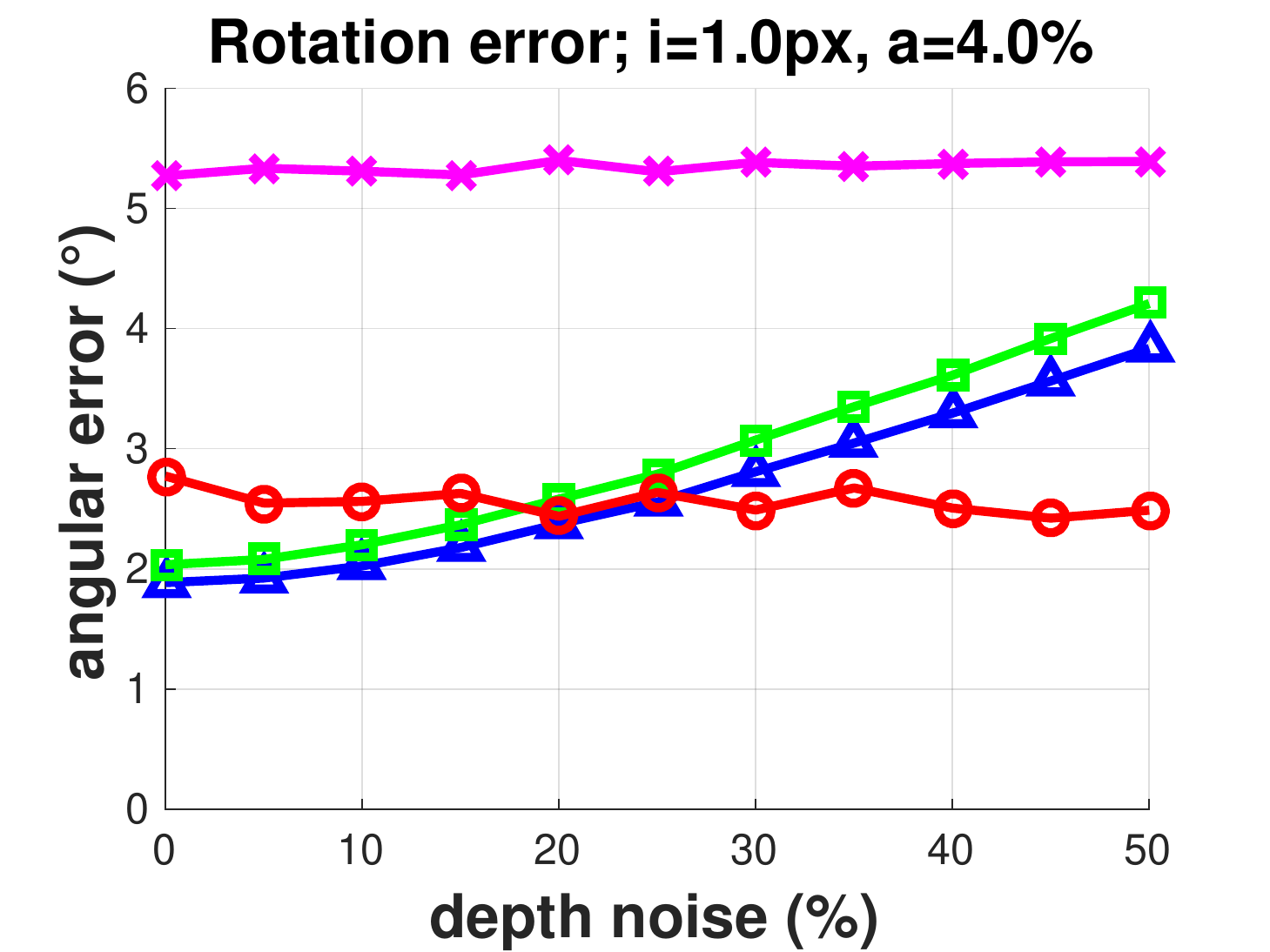}
        \label{fig:noisy_b}
	\end{subfigblock}\hfill
	\begin{subfigblock}{0.30\linewidth}
	    \includegraphics[trim={0.5cm 0 1cm 0},clip]{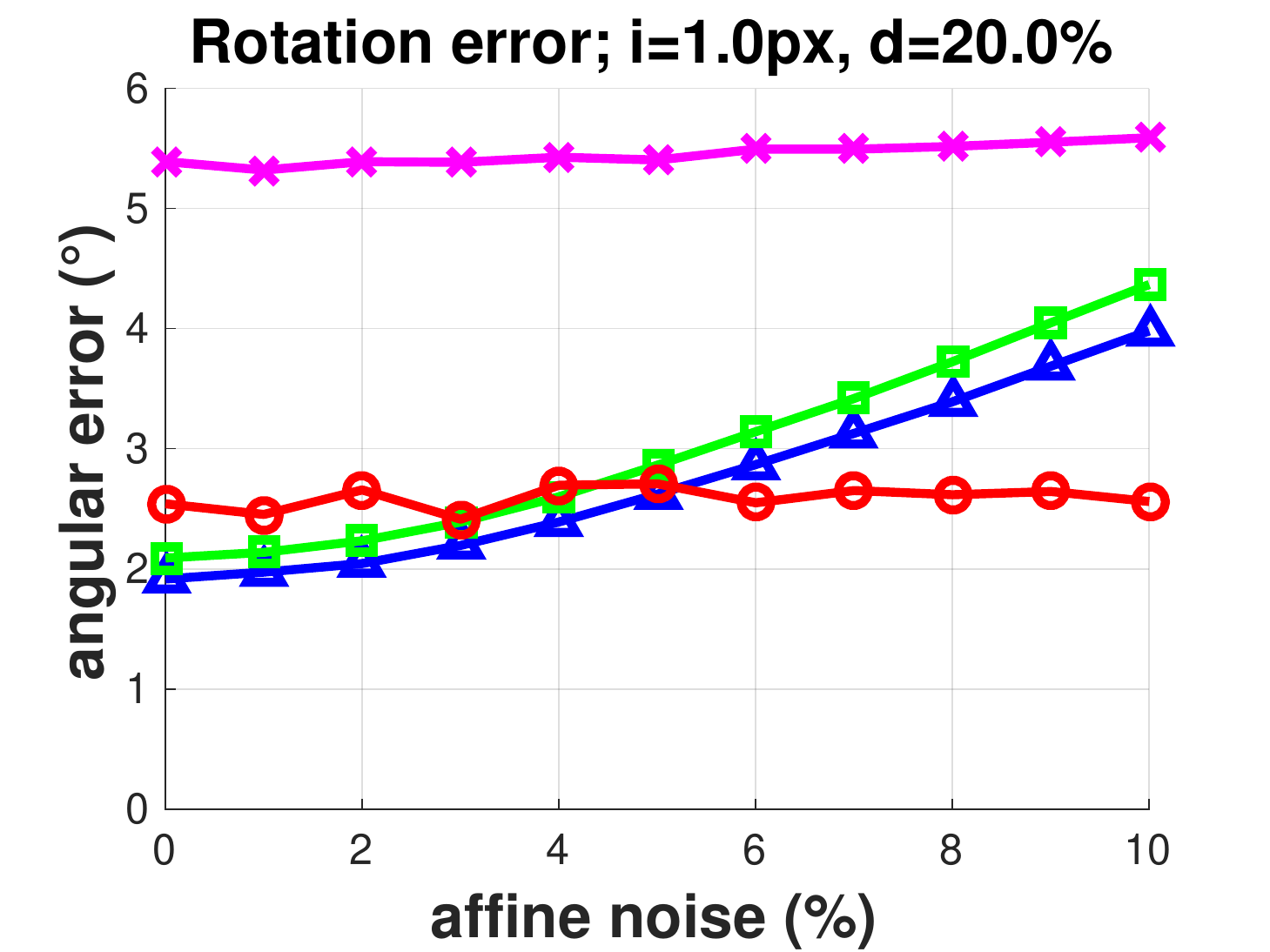}
        \label{fig:noisy_c}
	\end{subfigblock}\\

    \begin{subfigblock}{0.30\linewidth}
	    \includegraphics[trim={0.5cm 0 1cm 0},clip]{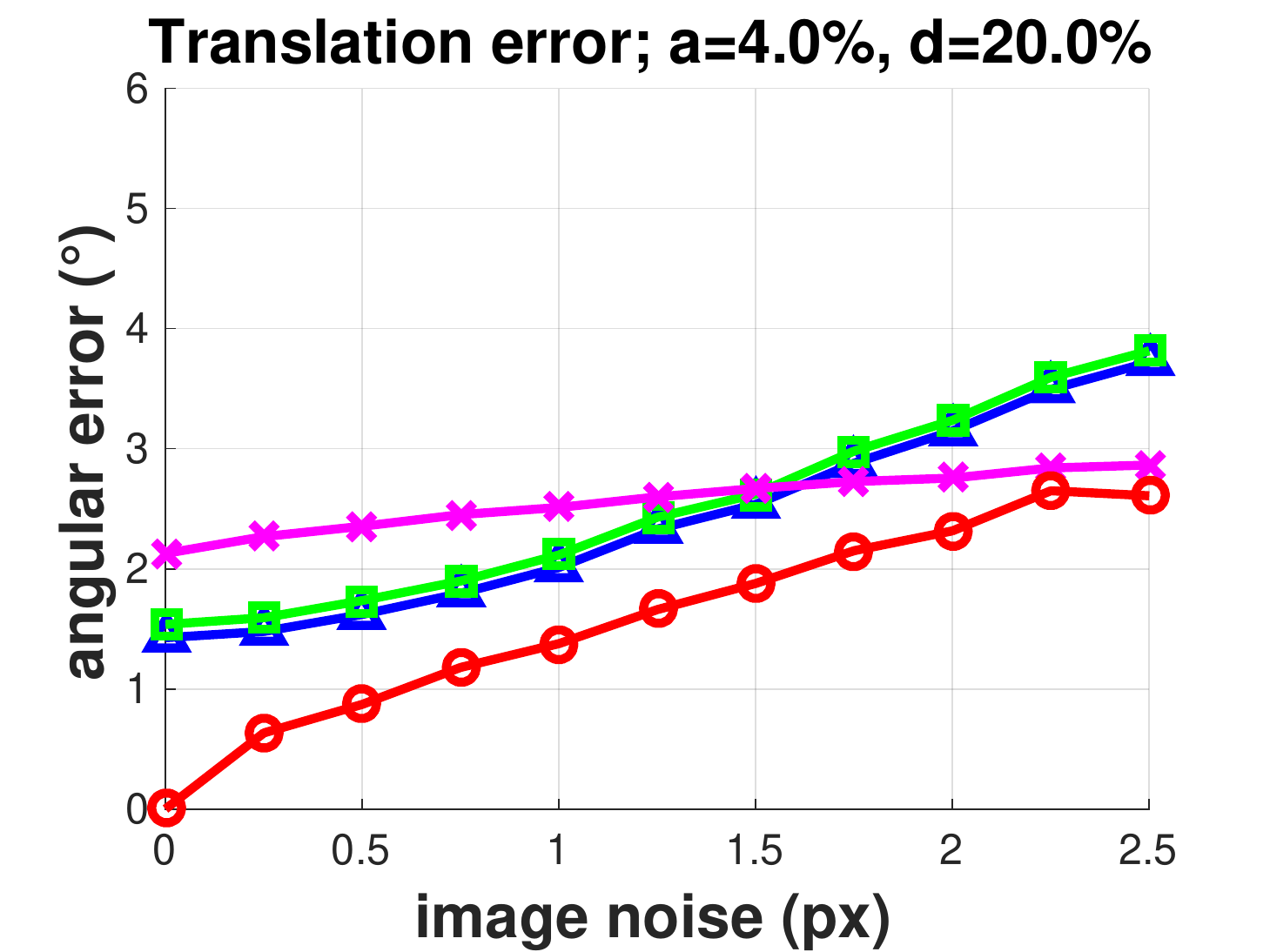}
        \label{fig:noisy_d}
	\end{subfigblock}\hfill
	\begin{subfigblock}{0.30\linewidth}
	    \includegraphics[trim={0.5cm 0 1cm 0},clip]{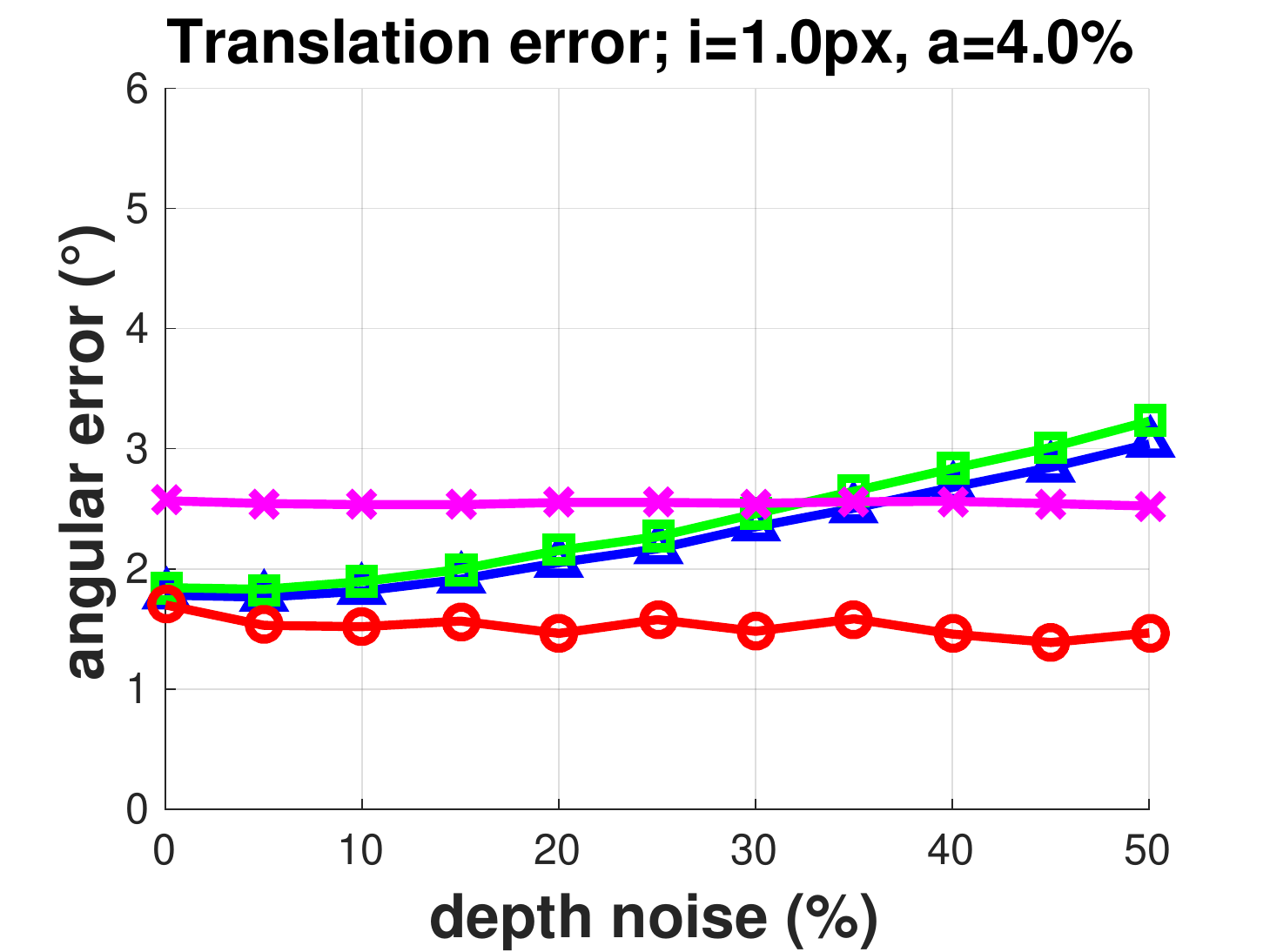}
        \label{fig:noisy_e}
	\end{subfigblock}\hfill
	\begin{subfigblock}{0.30\linewidth}
	    \includegraphics[trim={0.5cm 0 1cm 0},clip]{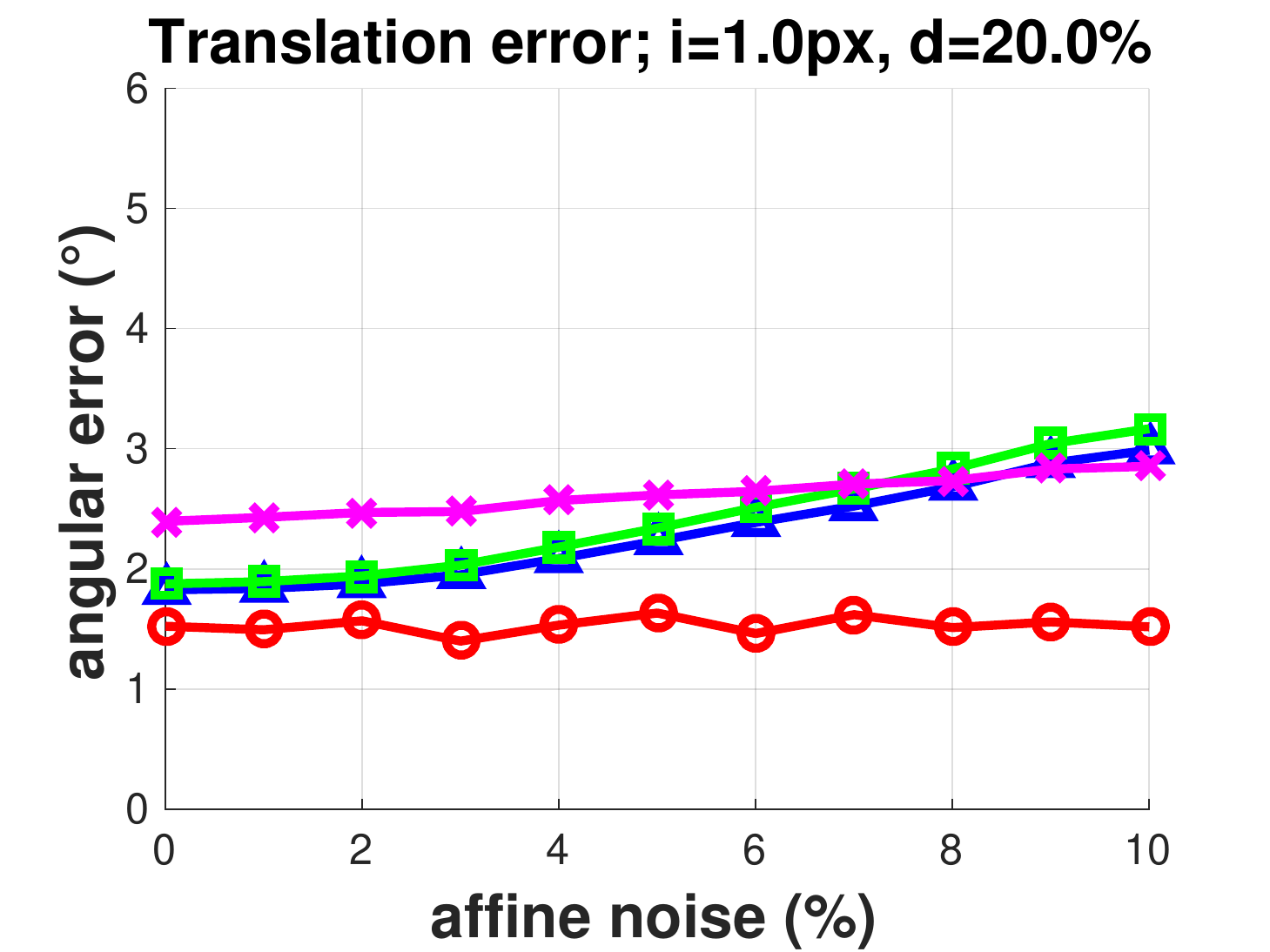}
        \label{fig:noisy_f}
	\end{subfigblock}
	\label{fig:noisy}
    \caption{Synthetic evaluation -- the effect of different levels of noise on various relative pose solvers. 
    Plots (a--f) compare the two proposed 1AC+D solvers, 2AC~\cite{barath2018efficient} and 5PT~\cite{stewenius2006recent}.
    In the 1st row (a--c), the rotation error is shown. 
    In the 2nd one (d--f) translation errors are plotted. 
    All errors are angular errors in degrees.
    In each column, different setups are shown, where we fixed two of the three sources of noise -- image, affine or depth noise -- to analyse the negative effect of the third as its level increases.}
\end{figure*}

\begin{figure}
    \centering
    \begin{namedsubfigblock}{0.62\linewidth}{1AC+D}
	    \includegraphics[width=1.0\textwidth]{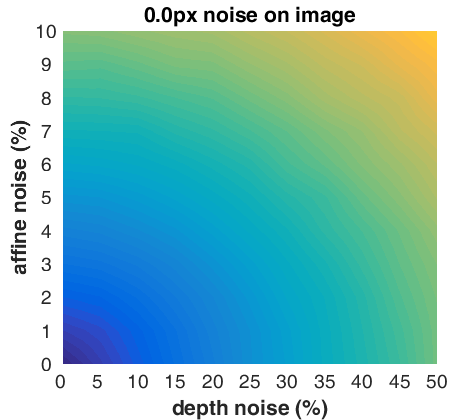}
        \label{fig:noisy_heatmap_1ACd_i}\hfill%
	    \includegraphics[width=1.0\textwidth]{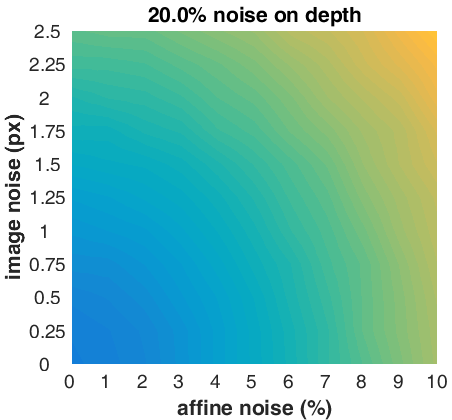}
	\end{namedsubfigblock}\hfill
	\begin{namedsubfigblock}{0.31\linewidth}{{\small 5PT}}
	    \includegraphics[width=1.0\textwidth]{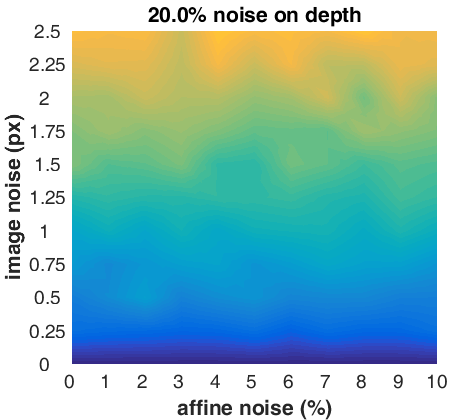}
        \label{fig:noisy_heatmap_5pt}
	\end{namedsubfigblock}\hfill
	\includegraphics[width=0.0392\textwidth]{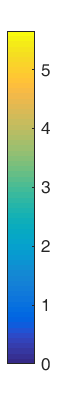}
    \caption{Synthetic evaluation -- the effect of image, affine and depth noise on the proposed 1AC+D and the point-based 5PT~\cite{stewenius2006recent} -- rotation errors, in degrees, displayed as heat-maps.
    As expected, 1AC+D is affected by all three noise types, as seen on (a). 
    Being a point-based relative pose solver, 5PT~\cite{stewenius2006recent} is affected only by the image noise (b).}
    \label{fig:my_label}
\end{figure}

\noindent \textbf{Noise study.}
We added noise to the \num{30000} instances of the synthetic setup. 
The estimated poses were evaluated to determine how sensitive the solvers are to the noise in the data, \ie, image coordinates, depth, and LAFs.
As expected, the noise in the depth or parameters of LAFs has no effect on the 5PT~\cite{stewenius2006recent} solver.
This can be seen in Fig.~\ref{fig:noisy_heatmap_5pt}, where 5PT~\cite{stewenius2006recent} is only affected by image noise. Increasing noise on either axes has a negative effect on the output of the proposed method as seen on Fig.~\ref{fig:noisy_heatmap_1ACd_i}.
The effect of noise on rotation and translation estimated by 1AC+D and 5PT~\cite{stewenius2006recent} are visualized on diagrams Fig.~\ref{fig:noisy_a}-(f).
The curves show that the effect of image noise -- on useful scales, such as \SI{2.5}{\px} -- in itself has a less significant effect on the degradation of rotation and translation estimates of the proposed method, compared to 5PT~\cite{stewenius2006recent}.
However, adding realistic scales of depth or affine noise, {\eg} Fig.~\ref{fig:noisy_a}, has an observable negative effect on the accuracy of the proposed ones.
Although 2AC~\cite{barath2018efficient} is unaffected by the noise on depth, it is moderately affected by noise on the affinities of LAFs. Its rotation estimates are worse than that of any other methods in the comparison.
The provided translation vectors are of acceptable quality.

Summarizing the synthetic evaluation, we can state that, on realistic scales of noise, the accuracy of the rotation and translation estimates of 5PT~\cite{stewenius2006recent} and both versions of 1AC+D are comparable. 2AC is the worst performer among all.

\subsection{Real-world evaluation}
\label{sec:experiments}

We tested the proposed solver on the 1DSfM dataset~\cite{wilson_eccv2014_1dsfm}\footnote{\url{http://www.cs.cornell.edu/projects/1dsfm/}}.
It consists of 13 scenes of landmarks with photos of varying sizes collected from the internet. 
1DSfM provides 2-view matches with epipolar geometries and a reference reconstruction from incremental SfM (computed with Bundler~\cite{snavely2006photo,snavely2008modeling}) for measuring error. 
We iterated through the provided 2-view matches, detected ACs~\cite{lowe1999object} using the VLFeat library~\cite{vedaldi08vlfeat}, applying the Difference-of-Gaussians algorithm combined with the affine shape adaptation procedure as proposed in~\cite{baumberg2000reliable}.
In our experiments, affine shape adaptation only had a small $\sim$\SI{10}{\percent} extra time demand, {\ie}, \SI[separate-uncertainty]{0.31 +- 0.25}{\second} per view, over regular feature extraction. This extra overhead is negligible, compared to the more time-consuming feature matching.
Matches were filtered by the standard ratio test~\cite{lowe1999object}. 
We did not consider image pairs in the evaluation with fewer than \num{20} corresponding features between them.
For the evaluation, we chose scenes Piccadilly, NYC Library, Vienna Cathedral, Madrid Metropolis, and Ellis Island.
To get monocular depth for each image, we applied the detector of Li~{\etal}~\cite{Li_2018_CVPR}. 
In total, the compared relative pose estimators were tested on a total of \num{110395} image pairs.

As a robust estimator, we chose the Graph-Cut RANSAC~\cite{barath2018graph} algorithm (GC-RANSAC) since it is state-of-the-art and has publicly available implementation\footnote{\url{https://github.com/danini/graph-cut-ransac}}.
In GC-RANSAC, and other locally optimized RANSACs, two different solvers are applied: (i) one for fitting to a minimal sample and (ii) one for fitting to a larger-than-minimal sample when improving the model parameters by fitting to all found inliers or in the local optimization step. 
For (i), the main objective is to solve the problem using as few points as possible since the overall wall-clock time is a function of the point number required for the estimation. 
For (ii), the goal is to estimate an accurate model from the provided set of points. 
In practice, step (ii) is applied rarely and, therefore, its processing time is not so crucial for achieving efficient robust estimation.
We used the normalized eight-point algorithm followed by a rank-2 projection to estimate the essential matrix from a larger-than-minimal sample.
We applied GC-RANSAC with a confidence set to $\num{0.99}$ and inlier-outlier threshold set to be the $\SI{0.05}{\percent}$ of the image diagonal size.
For the other parameters, the default values were used.

\begin{figure*}[t]
    \centering
\includegraphics[width=0.6\textwidth]{assets/noisebar_markers.pdf}\\
 	\begin{namedsubfigblock}{1.0\linewidth}{NYC Library}
 	    \includegraphics[width=0.31\textwidth]{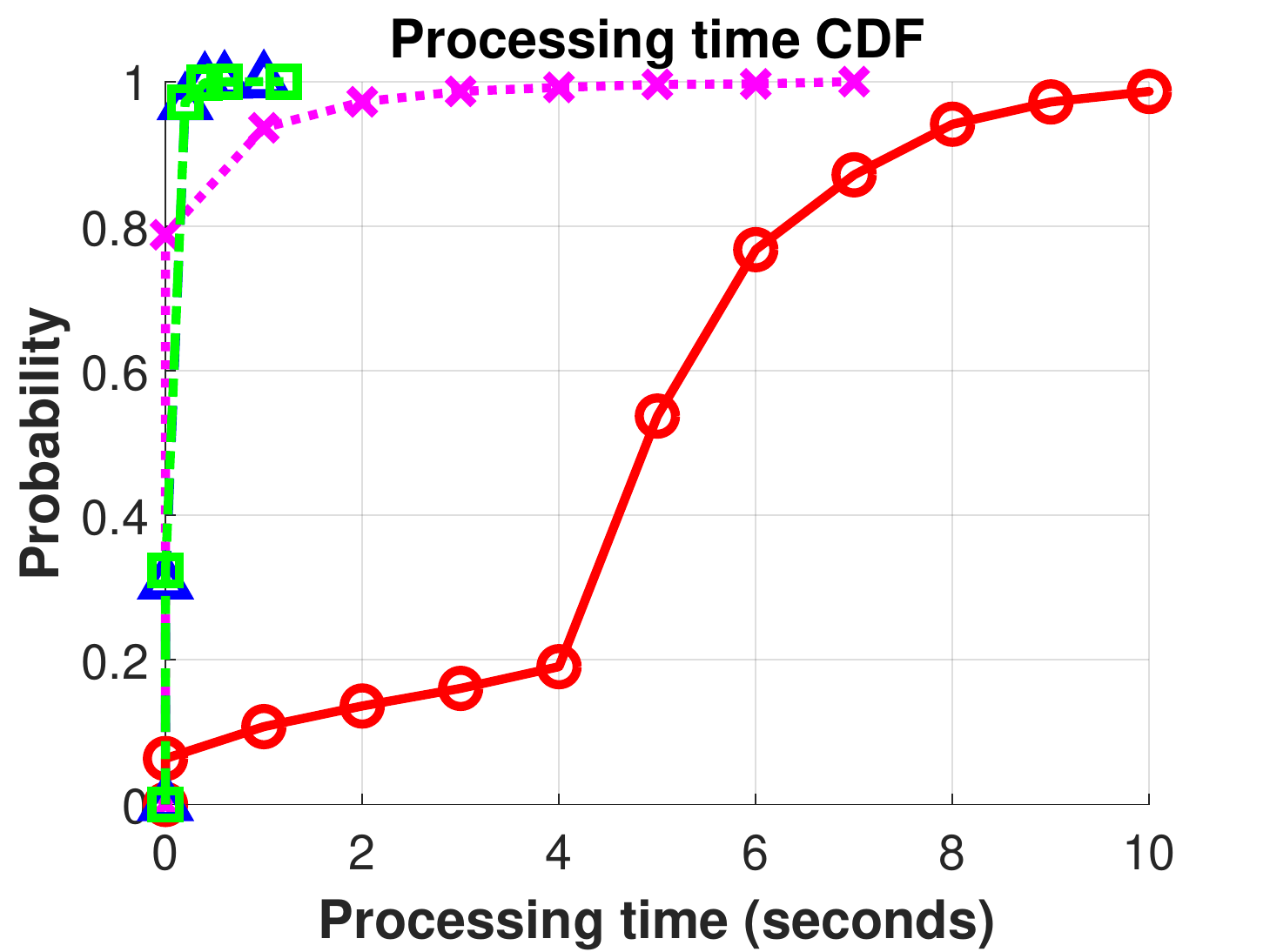}
 	    \includegraphics[width=0.31\textwidth]{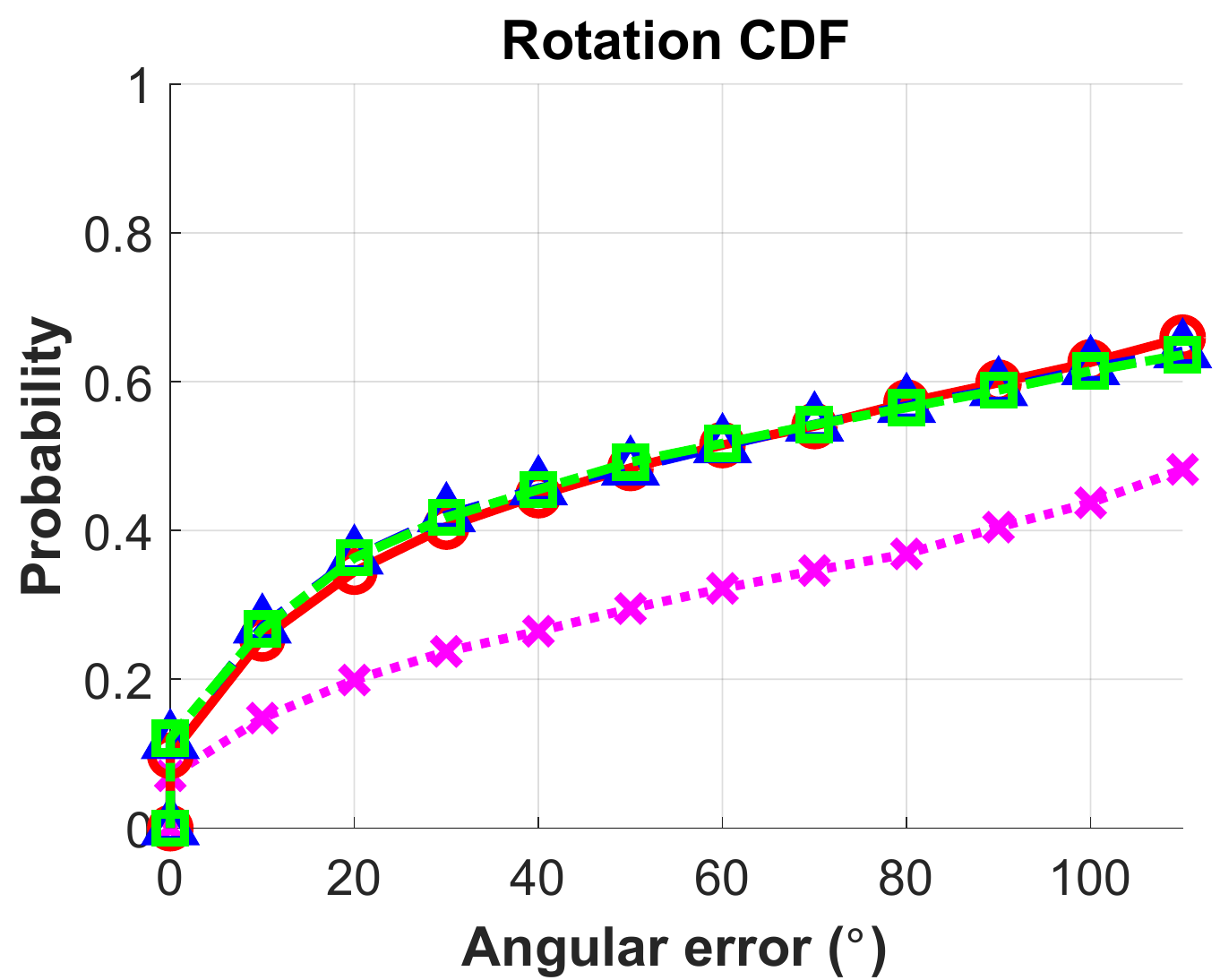}
 	    \includegraphics[width=0.31\textwidth]{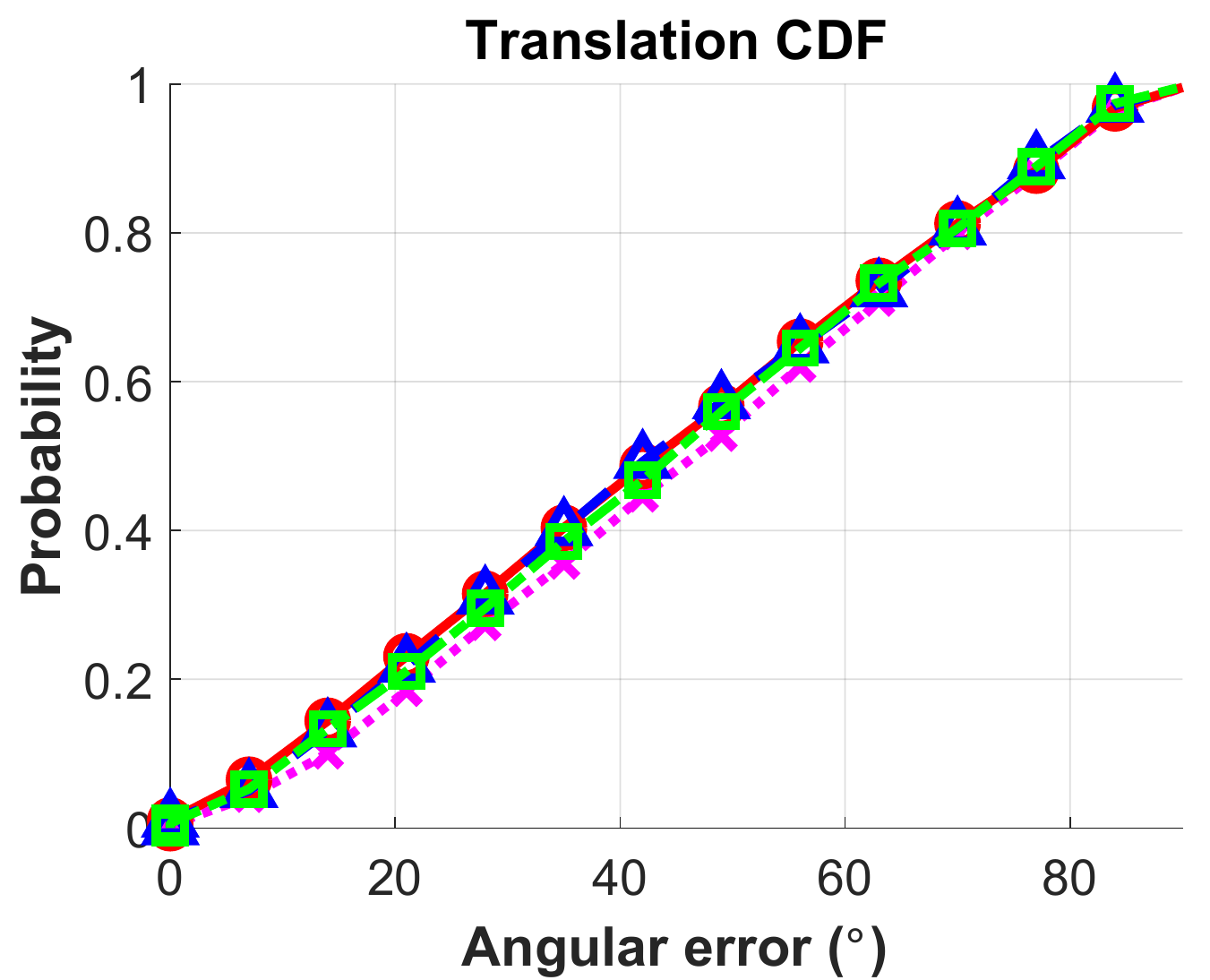}
 	\end{namedsubfigblock}
 	\begin{namedsubfigblock}{1.0\linewidth}{Madrid Metropolis}
 	    \includegraphics[width=0.31\textwidth]{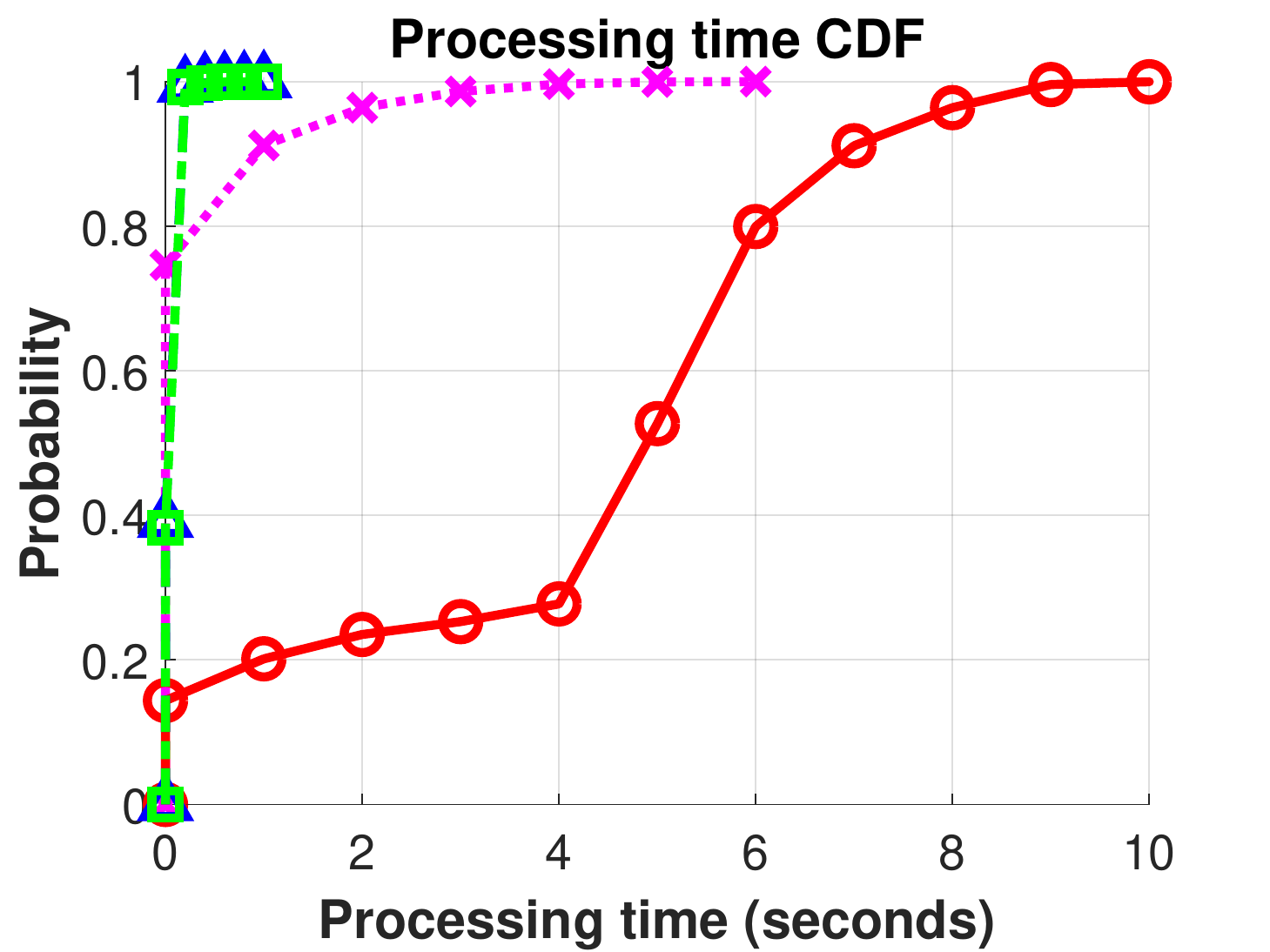}
 	    \includegraphics[width=0.31\textwidth]{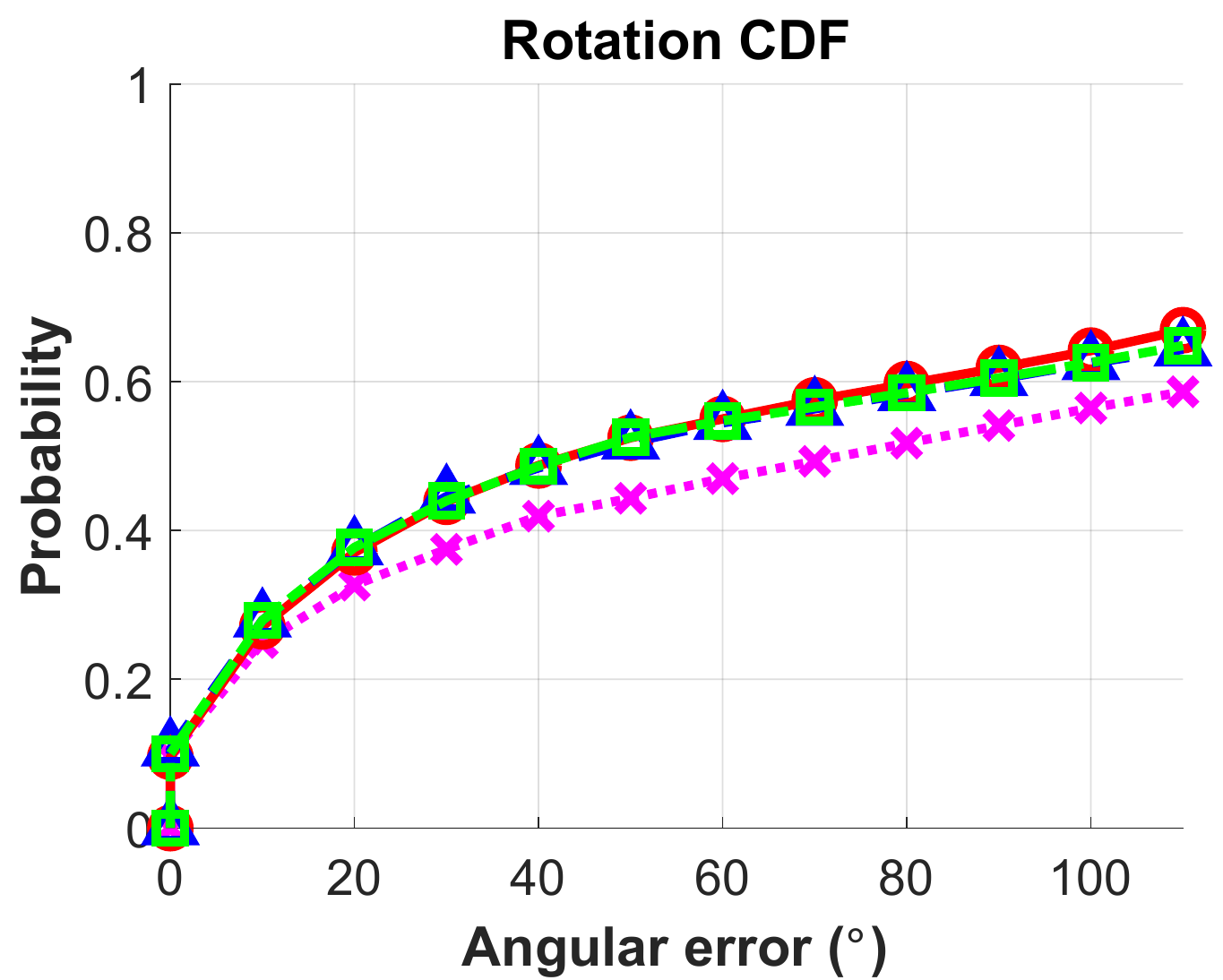}
 	    \includegraphics[width=0.31\textwidth]{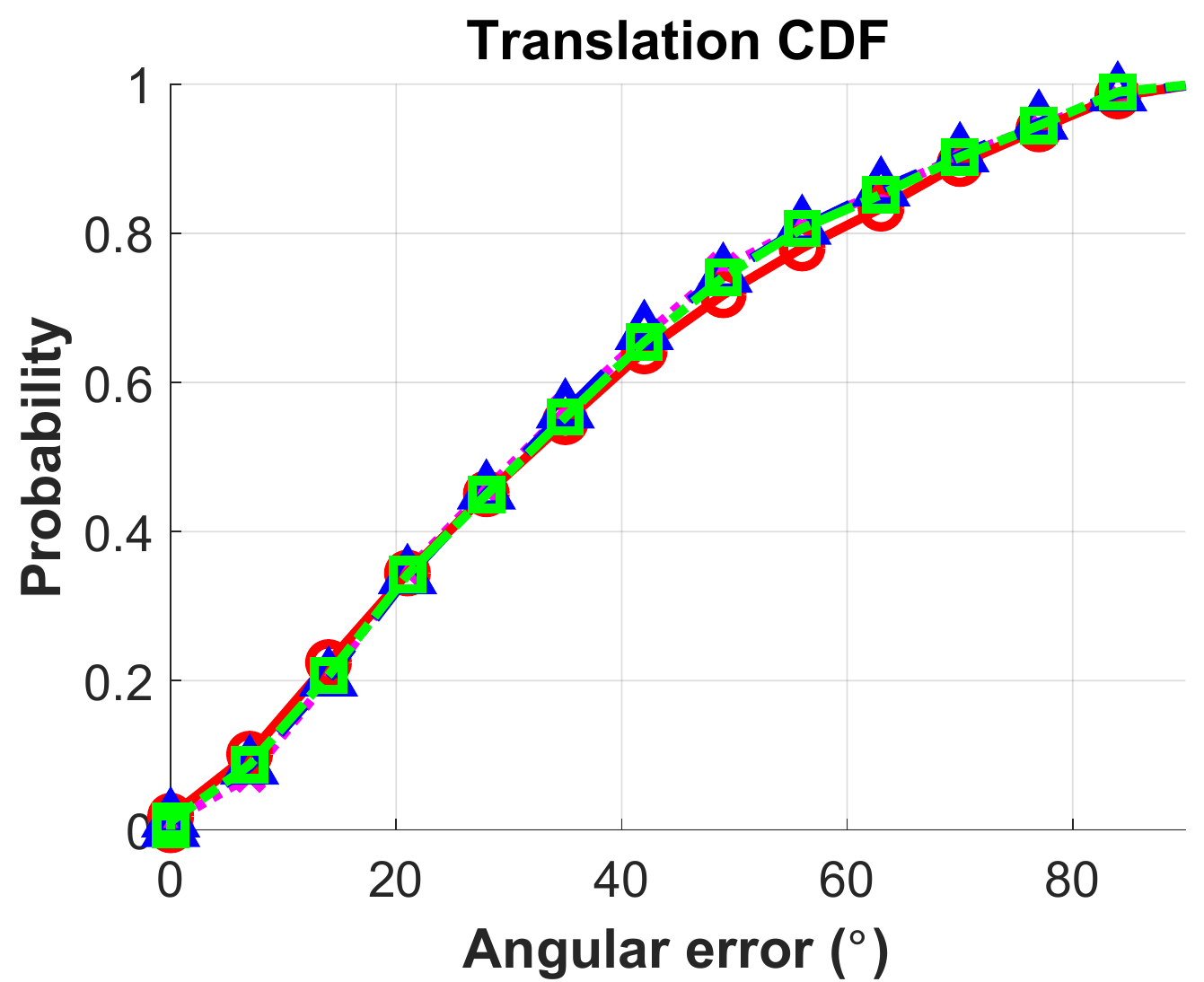}
 	\end{namedsubfigblock}
 	\begin{namedsubfigblock}{1.0\linewidth}{Vienna Cathedral}
 	    \includegraphics[width=0.31\textwidth]{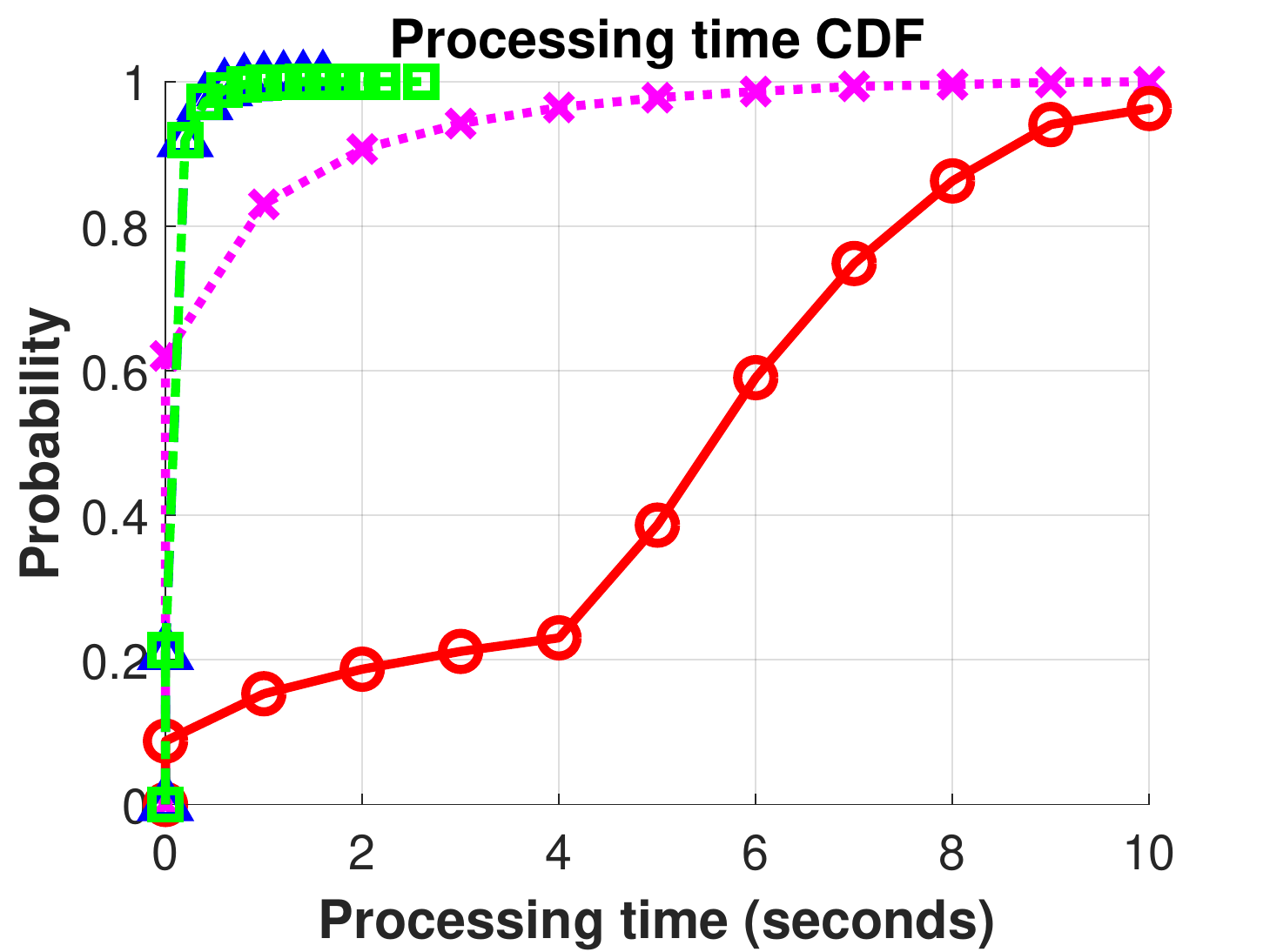}
 	    \includegraphics[width=0.31\textwidth]{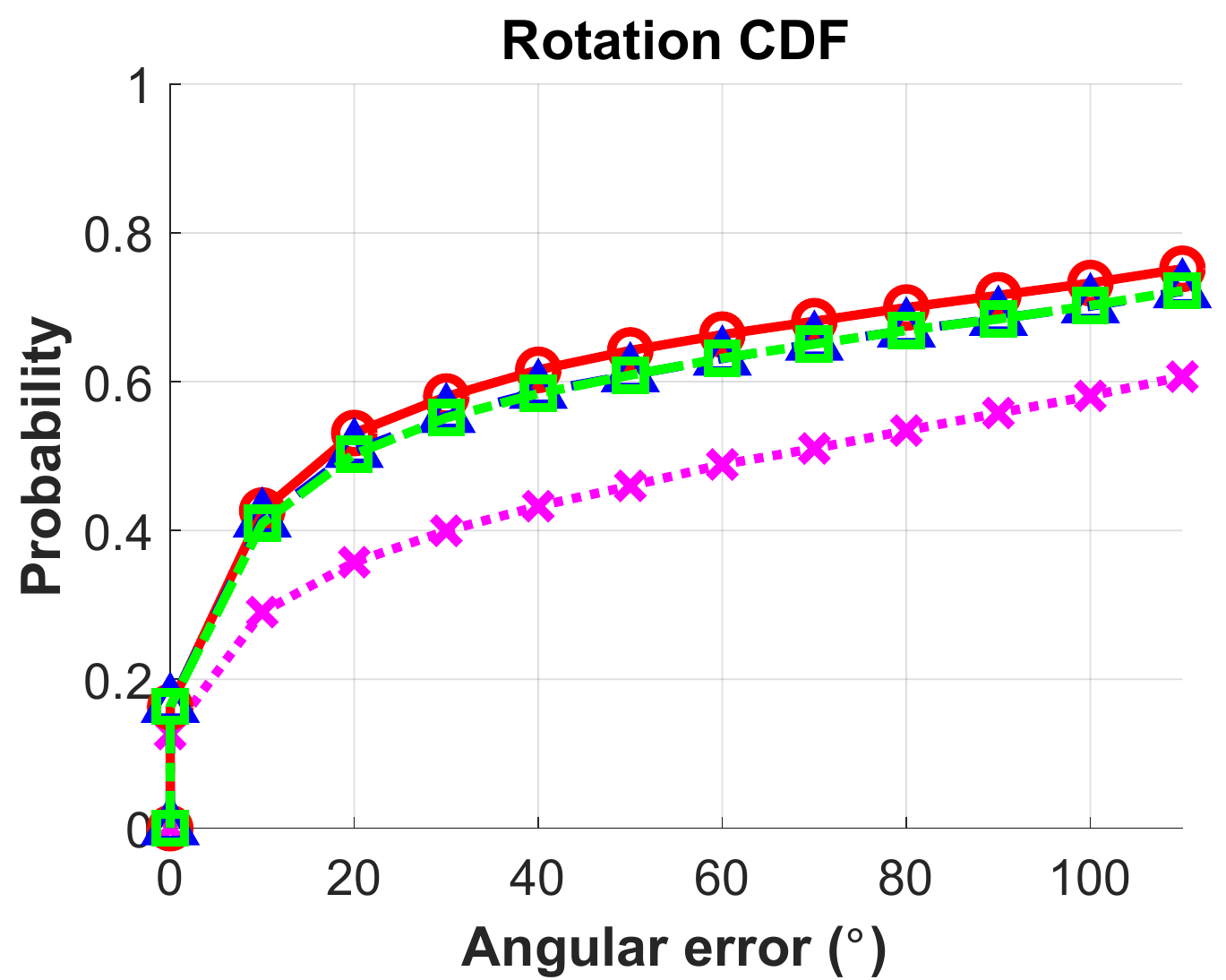}
 	    \includegraphics[width=0.31\textwidth]{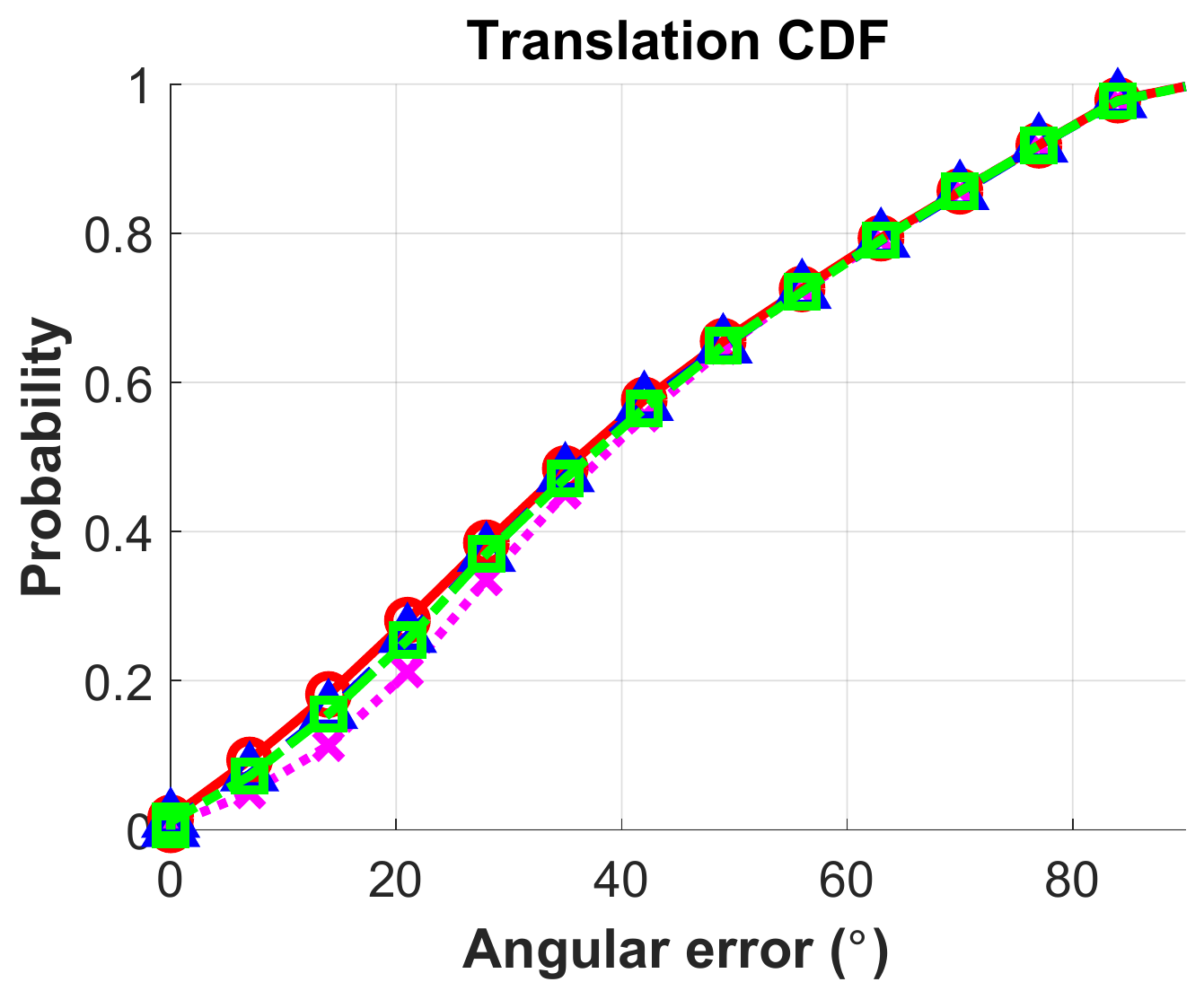}
 	\end{namedsubfigblock}
 	\begin{namedsubfigblock}{1.0\linewidth}{Ellis Island}
 	    \includegraphics[width=0.31\textwidth]{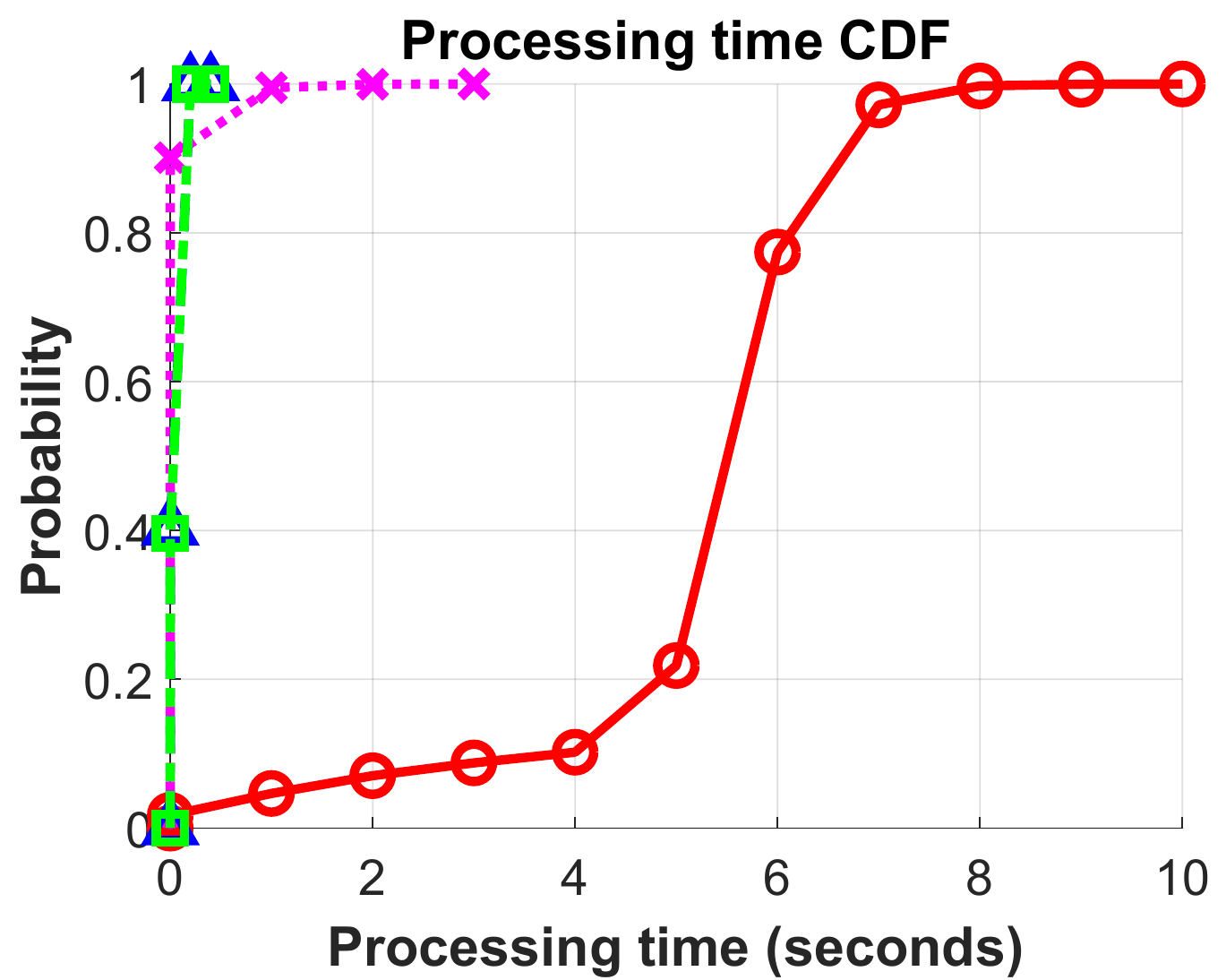}
 	    \includegraphics[width=0.31\textwidth]{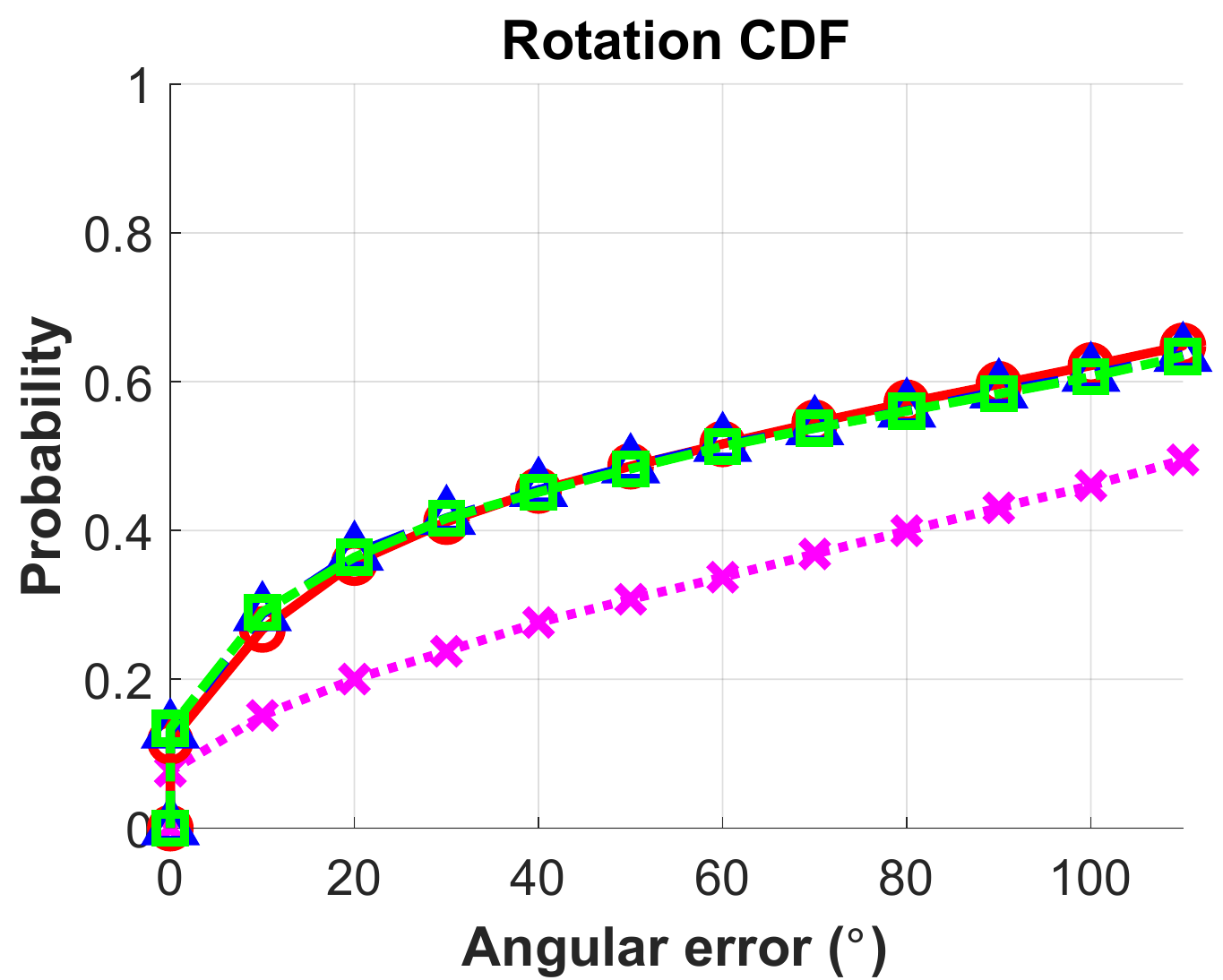}
 	    \includegraphics[width=0.31\textwidth]{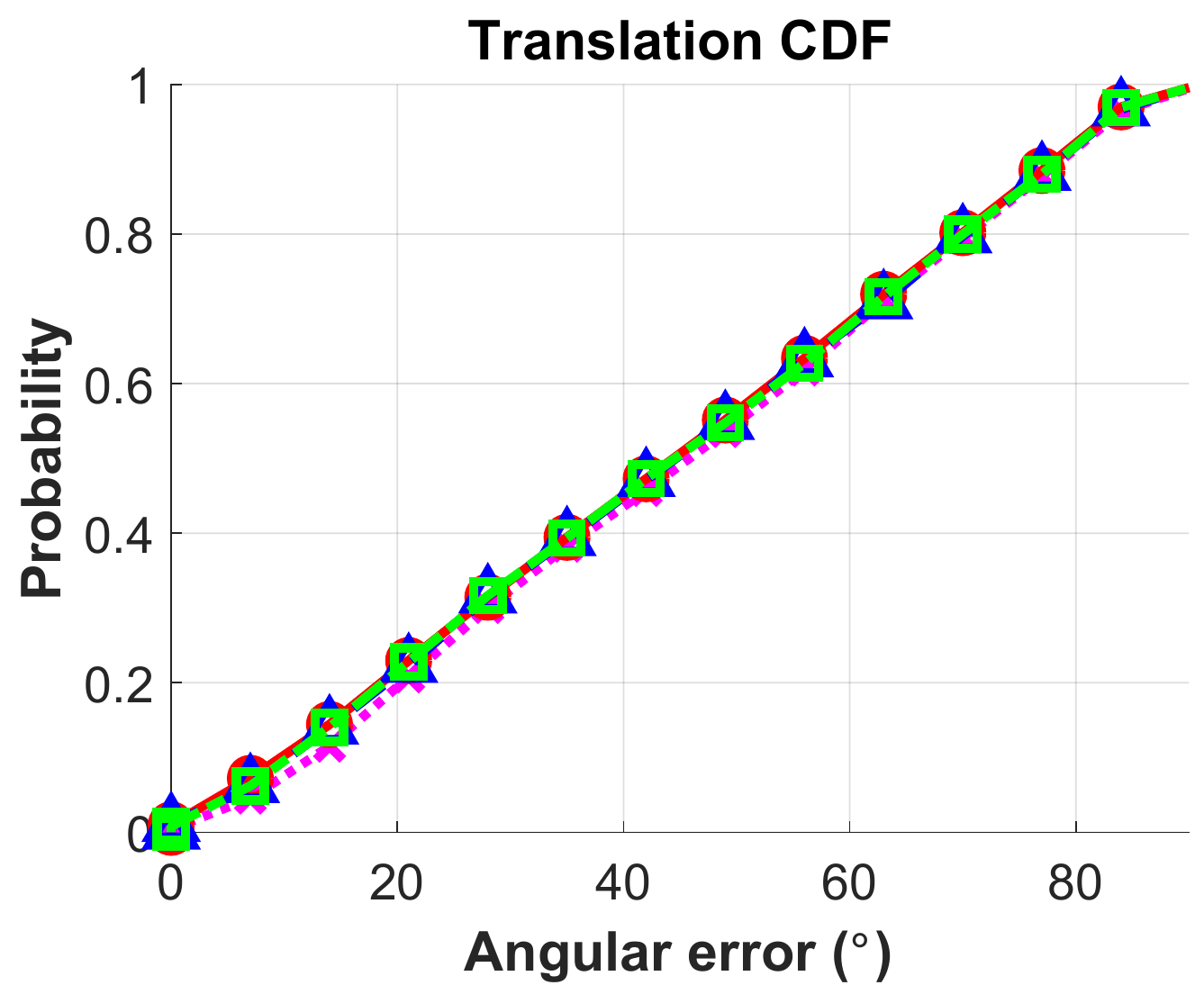}
 	\end{namedsubfigblock}
    \caption{Relative pose estimation on a total of \num{110395} image pairs from the 1DSfM dataset. 
    The cumulative distribution functions are shown for the processing time (in seconds), angular error of the estimated rotations and translations (in degrees).
    Being accurate or fast is interpreted as a curve close to the top-left corner.}
    \label{fig:cdf_errors}
\end{figure*}

Fig.~\ref{fig:cdf_errors} reports the cumulative distribution functions -- being accurate or fast is interpreted as a curve closer to the top-left corner -- calculated on all image pairs from each scene which was connected by an edge in the provided pose graph made by incremental SfM.
The left plot shows the processing time, in seconds, required for the full robust estimation.
The pose estimation is more than an order of magnitude faster when using the 1AC-D solver than the 5PT algorithm.
The right two plots show the rotation and translation errors calculated using the reference reconstructions of the 1DSfM dataset~\cite{wilson_eccv2014_1dsfm}.
1AC-D leads to accuracy similar to that of the 5PT algorithm, both in terms of rotation and translation.

Note that, in practice, multiple solvers are used for creating the initial pose graph, considering homography, fundamental and essential matrix estimation simultaneously, \eg, by QDEGSAC~\cite{frahm2006ransac}, to improve the accuracy and avoid degenerate configurations. 
It is nevertheless out of the scope of this paper, to include 1AC+D in state-of-the-art SfM pipelines.
However, in the next section we show as a proof-of-concept that the proposed solver can be used within global SfM pipelines and leads to similar accuracy as the widely-used 5PT method~\cite{stewenius2006recent}. 

\noindent \textbf{Applying global SfM algorithm.}
Once relative poses are estimated for camera pairs of a given dataset, along with the inlier correspondences, they are fed to the Theia library~\cite{theia-manual} that performs global SfM~\cite{chatterjee2013efficient,wilson_eccv2014_1dsfm} using its internal implementation.
That is, feature extraction, image matching and relative pose estimation were performed by our implementation either using the 1AC+D or the 5PT~\cite{stewenius2006recent} solvers, as described above.
The key steps of global SfM are robust orientation estimation, proposed by Chatterjee {\etal}~\cite{chatterjee2013efficient}, followed by robust nonlinear position optimization by Wilson {\etal}~\cite{wilson_eccv2014_1dsfm}.
The estimation of global rotations and positions enables triangulating 3D points, and the reconstruction is finalized by the bundle adjustment of camera parameters and points.

Table~\ref{tab:1DSfM} reports the results of Theia initialized by different solvers. 
There is no clear winner in terms of accuracy or run-time (of the reconstruction).
Both solvers perform similarly when used for initializing global structure-from-motion. 

\begin{table}[t]
\caption{The results of a global SfM~\cite{theia-manual} algorithm, on scenes from the 1DSfM dataset~\cite{wilson_eccv2014_1dsfm}, initialized with pose-graphs generated by the  5PT~\cite{stewenius2006recent} and 1AC+D solvers.
The reported properties are: the scene from the 1DSfM dataset~\cite{wilson_eccv2014_1dsfm} (1st column), relative pose solver (2nd), total runtime of the global SfM procedure given an initial pose-graph (3rd), rotation error of the reconstructed global poses in degrees (4th), position error in meters (5th) and focal length errors (6th).%
}%
\label{tab:1DSfM}
\resizebox{1.0\textwidth}{!}{%
\setlength\aboverulesep{0pt}\setlength\belowrulesep{0pt}%
\begin{tabular}{cc S[detect-weight,table-format=2.1] | S[detect-weight,table-format=1.1] S[detect-weight,table-format=1.1] S[detect-weight,table-format=2.1] | S[detect-weight,table-format=2.1] S[detect-weight,table-format=1.1] S[detect-weight,table-format=2.1] |S[detect-weight,table-format=1.1] S[detect-weight,table-format=1.1] S[detect-weight,table-format=1.1]}
\toprule
\rowcolor{black!10} 
 &
   &
   &
  \multicolumn{3}{c|}{\cellcolor{black!10}orientation (\SI{}{\degree})} &
  \multicolumn{3}{c|}{\cellcolor{black!10}position (m)} &
  \multicolumn{3}{c}{\cellcolor{black!10}focal len.\ ($\times \num{e-2}$)} \\
\hline
\rowcolor{black!10}
{scene}                        & {solver}    & {runtime (\SI{}{\second})} & {AVG} & {MED} & {STD}  & {AVG} & {MED} & {STD} & {AVG} & {MED} & {STD}   \\ \midrule
\multirow{2}{*}{Piccadilly}
& 1AC+D & 51,2 & \bfseries 6,1 & 4,3 & \bfseries 8,0 & 6,5 & \bfseries 3,4 & \bfseries 7,7 & \bfseries 2,3 & \bfseries 1.6 & \bfseries 3,2\\
& 5PT &  \bfseries 48,4 & 6,8 & \bfseries 2,5 & 10,1 & \bfseries 4,8 & 3,5 & 7,8 & 2,4 &  1,7 & 3,3\\ \midrule
\multirow{2}{*}{NYC Library}
& 1AC+D & 10,4 & 6,0 & 2,0 & \bfseries 6,3 & \bfseries 3,4 & \bfseries 3,6 & \bfseries 3,3 & \bfseries 2,7 & \bfseries 1,5 & \bfseries 4,1\\ 
& 5PT & \bfseries 5,9  & \bfseries 5,9 & \bfseries 1,9 & 6,7 & 4,2 & 3,7 & 5,5 & 2,9 & \bfseries 1,5 & 4,6\\ \midrule
\multirow{2}{*}{Vienna Cathedral}
& 1AC+D & \bfseries 56,5 & \bfseries 4,5 & \bfseries 1,7 & \bfseries 11,6 & \bfseries 6,1 & \bfseries 1,4 & \bfseries 10,3 & \bfseries 2,3 &  \bfseries 1,3 & \bfseries 3,7\\
& 5PT & 81,4 & 4,6 & 2,4 & 12,1 & 8,4 & 1,6 & 12,5 & \bfseries 2,3 & \bfseries 1,3 & 4,0\\ \midrule
\multirow{2}{*}{Madrid Metropolis}
& 1AC-D & \bfseries 14,6 & \bfseries 5,2 & \bfseries 2,4 & \bfseries 6,1 & \bfseries 13,6 & \bfseries 1,3 & \bfseries 15,5 & \bfseries 1,2 & \bfseries 0,5 & 2,4\\
& 5PT & 20,5 & 6,9 & 4,7 & 7,2 & 18,3 & 3,6 & 21,1 & \bfseries 1,2 & \bfseries 0,5 & \bfseries 2,3\\ \midrule
\multirow{2}{*}{Ellis Island}
& 1AC-D & 9,0 & 4,4 & \bfseries 5,6 & 7,2 &\bfseries  11,2 & 2,6 & \bfseries 11,2 & \bfseries 1,6 & 1,1 & \bfseries 1,7\\
& 5PT & \bfseries 5,7 & \bfseries 3,1 & 6,0 & \bfseries 3,9 & 11,8 & \bfseries 1,9 & 12,0 & \bfseries 1,6 & \bfseries 1,0 & \bfseries 1,7\\
\bottomrule
\end{tabular}
}%
\end{table}

\section{Conclusions}

In this paper, we propose a new approach for combining deep-learned non-metric monocular depth with affine correspondences (ACs) to estimate the relative pose of two calibrated cameras from a single correspondence. 
To the best of our knowledge, this is the first solution to the general relative camera pose estimation problem, from a single correspondence.
Two new general constraints are derived interpreting the relationship of camera pose, affine correspondences and relative depth.
Since the proposed solver requires a single correspondence, robust estimation becomes significantly faster, compared to traditional techniques, with speed depending linearly on the number of correspondences.

The proposed 1AC+D solver is tested both on synthetic data and on \num{110395} publicly available real image pairs from the 1DSfM dataset. 
It leads to an accuracy similar to traditional approaches while being significantly faster.
When solving large-scale problems, {\eg}, pose-graph initialization for Structure-from-Motion (SfM) pipelines, the overhead of obtaining affine correspondences and monocular depth is negligible compared to the speed-up gain in the pairwise geometric verification.
As a proof-of-concept, it is demonstrated on scenes from the 1DSfM dataset, via using a state-of-the-art global SfM algorithm, that acquiring the initial pose-graph by the proposed method leads to reconstruction of similar accuracy to the commonly used five-point solver.

\paragraph{Acknowledgement}
Supported by 
Exploring the Mathematical Foundations of Artificial Intelligence (2018-1.2.1-NKP-00008),
‘Intensification of the activities of HU-MATHS-IN—Hungarian Service Network of Mathematics for Industry and Innovation’ under grant number EFOP-3.6.2-16-2017-00015,
the Ministry of Education OP VVV project CZ.02.1.01/0.0/0.0/16 019/0000765 Research Center for Informatics, and 
the Czech Science Foundation grant GA18-05360S. 

\clearpage
\bibliographystyle{splncs04}
\bibliography{bibliography/header,bibliography/bib}

\chapter*{Appendix}

\section*{Validation of 1AC+D in the 1-point RANSAC scheme}

The proposed 1AC+D solver estimates the camera pose using a single correspondence and, thus, it can be applied in the 1-point RANSAC scheme.
In our proposal, it is also combined with the local optimization of  GC-RANSAC~\cite{barath2018graph}.
This supplementary material contains a brief evaluation of what inlier/outlier ratios are expected when using deep-learned depth prior~\cite{Li_2018_CVPR}, combined with affine correspondences~\cite{vedaldi08vlfeat}.

\subsubsection*{Applicability of local optimization.}
For this validation, it is interesting to see how many of the obtained correspondences lead to a sufficient number of inliers, when the proposed 1AC+D method is applied in the 1-point RANSAC scheme.
We consider an inlier number sufficient if the five-point algorithm, considering only the point coordinates from the correspondences, is applicable. 
When enough inliers are obtained, local optimization can further refine the inlier set while proposing an improved estimate of the camera pose.
Therefore, if at least $6$ inlier correspondences are found, the local optimization is applicable. 

We took all correspondences of all image pairs from two scenes -- \emph{i.e.} `Alamo' and `Madrid Metropolis' -- of the 1DSfM~\cite{wilson_eccv2014_1dsfm} dataset. To each image pair, we applied the proposed 1AC+D solver on each detected correspondence to determine the relative pose and measured the percentage of its inliers.
We took the average over all correspondences of all image pairs and found that
in scene `Madrid Metripolis', $\sim$\SI{25.1}{\percent} of the correspondences lead to an inlier set of at least 6 elements, 
while in `Alamo' $\sim$\SI{31.3}{\percent} of the correspondences lead to an inlier set of at least 6 elements.
In these cases local optimization is applicable to polish the pose parameters on a larger-than-minimal inlier set.

\subsubsection*{Applicability of 1-point RANSAC.}
The 1-point RANSAC scheme can handle particularly low number of inliers. Thus, to see the applicability of this scheme, we measured what percentage of correspondences lead to accurate pose estimation, when applying 1AC+D on them.

\begin{figure}[t]
    \centering
    \includegraphics[width=0.48\textwidth]{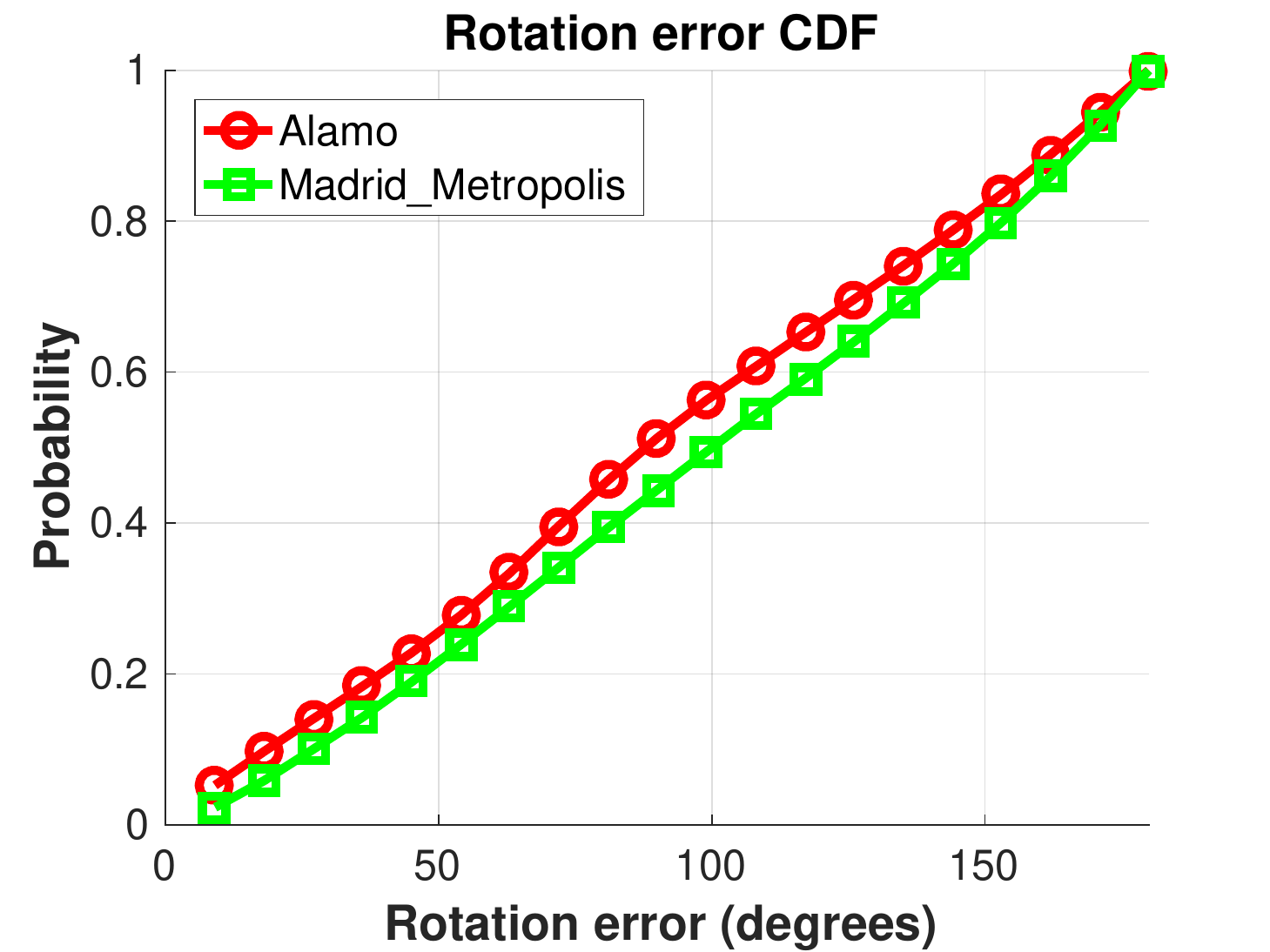}
    \includegraphics[width=0.48\textwidth]{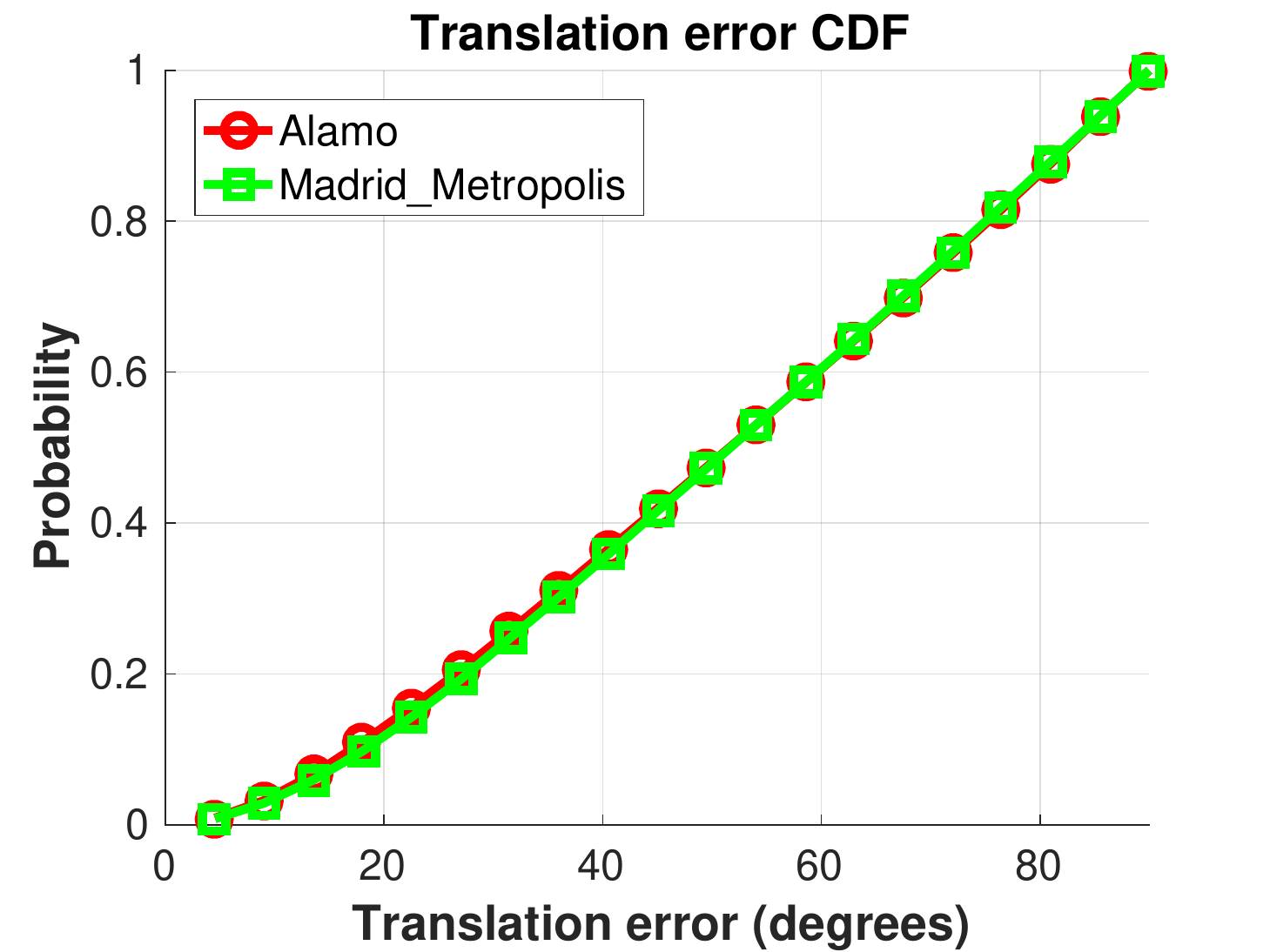}
    \caption{The CDFs of rotation and translation errors obtained using 1AC+D on all correspondences of all image pairs from two scenes of the 1DSfM~\cite{wilson_eccv2014_1dsfm} dataset. Being more accurate is interpreted as a curve close to the top-left corner.}
    \label{fig:cdf_errors_allcorresp}
\end{figure}
Fig.~\ref{fig:cdf_errors_allcorresp} shows the cumulative distribution functions (CDFs) of rotation and translation estimation errors for the `Alamo' and `Madrid Metropolis' scenes of the 1DSfM~\cite{wilson_eccv2014_1dsfm} dataset.
It is visible from the CDFs, that a fairly low percentage of correspondences lead to accurate pose estimates.
In fact, the percentage of correspondences that simultaneously yield low rotation and translation errors ($20^\circ$ and $15^\circ$, respectively) is even lower, about \SI{2.7}{\percent} and \SI{4}{\percent} for the two scenes.

Due to the fact that the proposed 1AC+D solver require only a single correspondence, such a low inlier ratio can still be handled and the accurate pose obtained. 
Given a confidence of $0.99$ and an inlier ratio of \SI{4}{\percent}, 1-point RANSAC would only need to perform \num{112} iterations. That is equivalent to only about \num{14896} FLOPs when using 1AC+D, not considering the few additional local optimization steps.
Consequently, even though the bad quality of the relative depth, the proposed solver is able to recover the sought pose parameters. 
For more details on the theoretical number of RANSAC iterations and the performance of our proposed method, the Reader is referred to Fig.~2 and Table~1 of our paper.

\subsubsection*{Summary.}
This brief evaluation suggests that, given the noisy correspondences obtained using the deep-learned depth priors and extracted affine features, 1AC+D produces a fairly low percentage of true inliers.
Note that given higher-quality data, {\eg} when using the metric depth from a depth camera, the above numbers are only expected to improve.
Even though, the proposed 1-point RANSAC scheme easily handles such a high outlier ratio, resulting in low number of iterations and accurate results.

\end{document}